\documentclass[10pt,journal,compsoc]{IEEEtran}

\usepackage[nocompress]{cite}

\usepackage[margin=0pt,font=small,labelfont=bf,labelsep=endash,tableposition=top]{caption}
\usepackage{epsfig}
\usepackage{graphicx}
\usepackage{amsmath}
\usepackage{amssymb}

\usepackage[accsupp]{axessibility}  %

\usepackage{graphicx}
\usepackage{amsmath}
\usepackage{amssymb}
\usepackage{booktabs}

\usepackage{listings}
\usepackage{hyperref}

\usepackage{multirow}

\usepackage{enumitem}
\usepackage{tabularx}
\usepackage{multirow}
\usepackage{stfloats}
\usepackage{threeparttable}
\usepackage[usenames,dvipsnames]{xcolor}

\usepackage{bm}

\newlength\savedwidth

\definecolor{darkergreen}{RGB}{21, 152, 56}
\definecolor{red2}{RGB}{252, 54, 65}

\definecolor{blacktext}{RGB}{0, 0, 0}

\usepackage{xcolor}
\definecolor{ww_color}{RGB}{128, 255, 0}

\definecolor{blue3}{RGB}{20, 54, 254}
\definecolor{blue2}{RGB}{20, 54, 254}

\usepackage{bbding}

\usepackage[american]{babel}
\usepackage{microtype}

\usepackage{subcaption}

\usepackage[capitalize]{cleveref}
\crefname{section}{Sec.}{Secs.}
\Crefname{section}{Section}{Sections}
\crefname{table}{Tab.}{Tabs.}
\Crefname{table}{Table}{Tables}
\crefname{figure}{Fig.}{Figs.}
\Crefname{figure}{Figure}{Figures.}

\usepackage{xspace}
\usepackage{adjustbox}
\usepackage{threeparttable}
\usepackage{multirow}
\usepackage{makecell}
\usepackage{siunitx}
\usepackage{tcolorbox}

\usepackage{adjustbox}
\usepackage{threeparttable}
\usepackage{xcolor}
\usepackage{multirow}
\usepackage{makecell}
\usepackage{siunitx}
\usepackage{tcolorbox}
\usepackage{listings}
\usepackage{xcolor}
\usepackage{placeins}

\definecolor{codegreen}{rgb}{0,0.6,0}
\definecolor{codegray}{rgb}{0.5,0.5,0.5}
\definecolor{codepurple}{rgb}{0.58,0,0.82}
\definecolor{backcolour}{rgb}{0.95,0.95,0.92}

\definecolor{deepyellow}{RGB}{255, 215, 0}

\lstdefinestyle{mystyle}{
    backgroundcolor=\color{backcolour},   
    commentstyle=\color{codegreen},
    keywordstyle=\color{magenta},
    numberstyle=\tiny\color{codegray},
    stringstyle=\color{codepurple},
    basicstyle=\ttfamily\footnotesize,
    breakatwhitespace=false,         
    breaklines=true,                 
    captionpos=b,                    
    keepspaces=true,                 
    numbers=left,                    
    numbersep=5pt,                  
    showspaces=false,                
    showstringspaces=false,
    showtabs=false,                  
    tabsize=2
}
\newcommand{\ourmodel}{{\textsc{OmniParser V2}}\xspace}
\newcommand{\pointsdecoder}{{token-router-based shared decoder}\xspace}

\newcommand\mypara[1]{\vspace{1.0mm}\noindent\textbf{#1}}
\newcommand\rankfirst[1]{\textbf{#1}}
\newcommand\ranksecond[1]{\underline{#1}}
\newcommand\token[1]{\texttt{\textless{#1}\textgreater}}

\usepackage{longtable}
\usepackage{amsmath}

\begin{document}

\title{OmniParser V2: Structured-Points-of-Thought for Unified Visual Text Parsing and Its Generality to Multimodal Large Language Models}

\author{
Wenwen Yu,
Zhibo Yang,
Jianqiang Wan,
Sibo Song,
Jun Tang,
Wenqing Cheng, \\
Yuliang Liu, IEEE Member,
Xiang Bai*, IEEE Fellow

\IEEEcompsocitemizethanks{
\IEEEcompsocthanksitem W. Yu is with the School of Information Science and Engineering, East China University of Science and Technology, Shanghai, 200237, China. (email: wenwenyu@ecust.edu.cn).
\IEEEcompsocthanksitem Y. Liu is with the School of Artificial Intelligence and Automation, Huazhong University of Science and Technology, Wuhan, 430074, China (email: ylliu@hust.edu.cn).
\IEEEcompsocthanksitem X. Bai is with the School of Software Engineering, Huazhong University of Science and Technology, Wuhan, 430074, China (email: xbai@hust.edu.cn).
\IEEEcompsocthanksitem W. Cheng is with the School of Electronic Information and Communications, Huazhong University of Science and Technology, Wuhan, 430074, China (email: chengwq@hust.edu.cn).
\IEEEcompsocthanksitem Z. Yang, J. Wan, S. Song, and J. Tang are with the Alibaba Group, Hangzhou, 310000, China (email: \{yangzhibo450, hustwjq, sibosongzju, tjbestehen\}@gmail.com).
}
\thanks{This work was supported by the National Natural Science Foundation of China (No.62225603, No.62206104), the National Key Research and Development Program (No.2022YFC2305102), and Alibaba Innovative Research (AIR) program.}
\thanks{Corresponding author: Xiang Bai.}
}

\IEEEtitleabstractindextext{
\begin{abstract}
Visually-situated text parsing (VsTP) has recently seen notable advancements, driven by the growing demand for automated document understanding and the emergence of large language models capable of processing document-based questions. While various methods have been proposed to tackle the complexities of VsTP, existing solutions often rely on task-specific architectures and objectives for individual tasks. This leads to modal isolation and complex workflows due to the diversified targets and heterogeneous schemas. In this paper, we introduce OmniParser V2, a universal model that unifies VsTP typical tasks, including text spotting, key information extraction, table recognition, and layout analysis, into a unified framework. Central to our approach is the proposed Structured-Points-of-Thought (SPOT) prompting schemas, which improves model performance across diverse scenarios by leveraging a unified encoder-decoder architecture, objective, and input\&output representation. SPOT eliminates the need for task-specific architectures and loss functions, significantly simplifying the processing pipeline. Our extensive evaluations across four tasks on eight different datasets show that OmniParser V2 achieves state-of-the-art or competitive results in VsTP. Additionally, we explore the integration of SPOT within a multimodal large language model structure, further enhancing visual text parsing capabilities on four tasks, thereby confirming the generality of SPOT prompting technique. The code is available at \href{https://github.com/AlibabaResearch/AdvancedLiterateMachinery}{https://github.com/AlibabaResearch/AdvancedLiterateMachinery}.

\end{abstract}

}

\maketitle

\

\IEEEraisesectionheading{\section{Introduction}
\label{sec:introduction}}

\begin{figure}[htbp]
    \centering 
    \centerline{\includegraphics[width=1.0\linewidth]{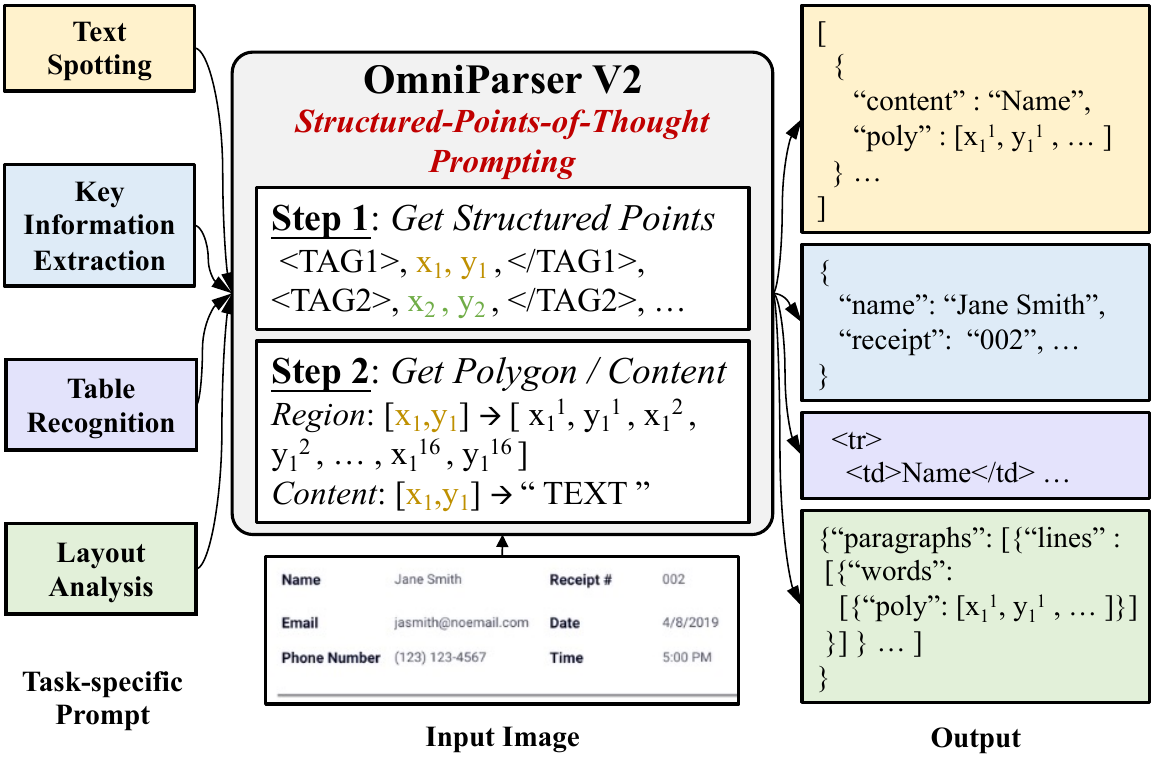}}
    \vspace{-2mm}
    \caption{\textbf{A task-agnostic architecture for visually-situated text parsing. } The proposed \ourmodel takes an image and a task-specific structured-points-of-thought prompting as input and generates structured text sequences tailored to the specified task, including text spotting, key information extraction, table recognition, and layout analysis.}
    \label{fig:pipeline}
    \vspace{-6mm}
\end{figure}

\IEEEPARstart{V}{i}sually-situated text parsing (VsTP) refers to the extraction of structured information from document images by jointly interpreting visual and textual elements within the text-rich image, such as text, tables, graphics, layout structures, and other visual entities, partly shown in~\cref{fig:pipeline}. With the exponential growth of text-related data and the rapid advancements in Large Language Models (LLM)~\cite{ChatGPT, GPT-4} and Multimodal Large Language Models (MLLM)~\cite{GPT-4V(ision)}, there has been recently a surge of research on the topic of VsTP~\cite{Chen2023PaLIXOS, li2023blip2, li2022relational, ye2023deepsolo}. Existing approaches can be broadly classified into two categories: generalist models~\cite{Chen2023PaLIXOS, li2023blip2} and specialist models~\cite{ ye2023deepsolo, long2021parsing, wang2021towards}.

Both generalist models and specialist models have limitations in handling multiple multimodal tasks that are closely interconnected in the domain of VsTP. Generalist models excel in their versatility and universality across domains, but often struggle with achieving high precision and interpretability. Their performances can be significantly constrained when an external Optical Character Recognition (OCR) engine is unavailable~\cite{Chen2023PaLIXOS}. Additionally, the prediction processes of such models are usually non-transparent, due to their black-box nature. In contrast, specialist models often achieve superior performance in their specific sub-tasks~\cite{long2021parsing, ye2023deepsolo}. However, when confronted with the requirement of multitasking, the pipeline will be usually more complex. Moreover, discrete specialist models inadvertently lead to modal isolation and limit in-depth understanding.

In recent years, there has been a growing trend toward unified models capable of performing multiple visually-situated text parsing tasks, as illustrated in ~\cref{table:unified}. While these models have demonstrated promising effectiveness, handling the diverse text structures and various relations in VsTP remains a significant challenge. Accordingly, tasks in visual document parsing can be categorized into: 1) Sequential text detection and recognition, 2) Table structure and content recognition, and 3) Visual entity extraction and localization. Developing a unified framework that effectively addresses these diverse tasks while maintaining high performance poses several challenges. First, incorporating task-specific heads~\cite{ye2023deepsolo}, adapters~\cite{Liu2021ABCNetVA, Liao2020MaskTV}, and formulations~\cite{long2021parsing, kim2022donut} can hinder achieving generality. Second, handling cross-dependencies between tasks is crucial. For instance, table recognition inherently involves text spotting. Third, a unified representation of tasks should consider both primary visual elements~(\textit{words, points, lines, cells}) and various types of relations~(\textit{the adjacency between characters, the linking between keys and values, and the alignment of table cells.}).

Along with this line of work, we introduce a unified paradigm for VsTP, named \textit{\textbf{OmniParser V2}}. This method employs a basic image encoder and a token-router-based shared decoder that is implemented by a simplified mixture-of-experts (MoE)~\cite{Shazeer2017OutrageouslyLN,NEURIPS2022_d46662aa} based Transformer decoder to reduce model size. By adopting a single architecture, standardizing modeling objective as well as output representation, \ourmodel seamlessly handles multiple typical VsTP tasks, including text spotting, key information extraction (KIE), table recognition (TR), and layout analysis, in a unified framework, as illustrated in~\cref{fig:pipeline}. To enhance performance and improve transparency, we propose Structured-Points-of-Thought (SPOT) prompting, a two-stage generation strategy. In the first stage, a \textit{structured points sequence} consisting of center points of text segments, along with task-related structural tokens, is generated via the token-router-based shared decoder, conditioned on the embeddings of the input image and task prompt. In the second stage, given each text center point obtained from the first stage as model prompting, both the \textit{polygon and content sequence} are predicted by the same token-router-based shared decoder. By leveraging the outputs of these two stages, OmniParser V2 can efficiently format task-specific results in a structured manner, enabling a unified and effective approach to VsTP.

The rationale behind the two-stage design is straightforward. The first stage generates center point sequences, which effectively represent word-level/line-level text instances while preserving complex structures encoded in markup languages such as JSON or HTML. The second stage can uniformly generate polygonal contours and recognition results in parallel across different VsTP tasks. An obvious advantage of the two-stage strategy of structured-points-of-thought prompting is its explicit decoupling of structured sequence learning, which significantly reduces sequence length and, in turn, simplifies the learning process. This reduction leads to higher performance and better generalization across diverse VsTP tasks. Additionally, we explore the incorporation of SPOT prompting into multimodal large language models (MLLMs) for VsTP, a domain where MLLMs traditionally underperformed~\cite{Liu2023OCRBenchOT,Yang2024CCOCRAC,fu2024ocrbenchv2}, observing substantial improvements in visual text parsing performance across four tasks, confirming the generality of the SPOT technique.

Our major contributions are as follows:
\begin{itemize}
    \item We propose \ourmodel, a unified framework for visually-situated text parsing, capable of simultaneously handling text spotting, key information extraction, table recognition, and layout analysis. The proposed SPOT prompting paradigm, as a core component of our framework, is applicable to both lightweight models and large multimodal models.
    \item We introduce a two-stage structured-points-of-thought prompting technique, where structured points sequence serves as intermediate results of the model. This paradigm enhances the models' ability to parse structural information, while improving interpretability.    
    \item We present a token-router-based shared decoder that effectively reduces model size. Additionally, we design two pre-training strategies, namely spatial-aware prompting and content-aware prompting, which enhance the \pointsdecoder for richer spatial and semantic representation learning in VsTP. 
    \item Experiments on standard benchmarks demonstrate that \ourmodel outperforms existing unified models across all four tasks while maintaining competitive performance against specialized, task-specific models.
    \item We investigate how integrating SPOT prompting into a multimodal large language model enhances visual text parsing performance, highlighting the broad applicability of the SPOT prompting technique. 
\end{itemize}

\begin{table}[t]
\centering
\caption{\textbf{Comparing the parsing capabilities achieved by different unified paradigms.} `TSR' and `TCR' denote Table Structure Recognition and Table Content Recognition respectively. To the best of our knowledge, \ourmodel is the first paradigm that accomplishes end-to-end visually-situated text parsing for text spotting, key information extraction, table recognition, and layout analysis.}
\setlength\tabcolsep{1.3pt}
\begin{adjustbox}{max width=0.49\textwidth}
\begin{tabular}{lcccc}
\toprule
\multirow{2}{*}{Methods}       & \multicolumn{4}{c}{Visually-situated Text Parsing}    \\ \cmidrule(l){2-5} 
                                & Text Spotting & KIE      & Table Recognition & Layout Analysis   \\ \midrule
Donut~\cite{kim2022donut}      &    $\times$        & E2E, w/o Loc.           &   $\times$  &   $\times$        \\
BROS~\cite{hong2022bros}       &     $\times$       & OCR-dependent        & TSR   &   $\times$      \\
DocReL~\cite{li2022relational} &     $\times$       & OCR-dependent        & TSR    &   $\times$     \\
UniDoc~\cite{feng2023unidoc}   & $\checkmark$        & E2E, w/o Loc.           &  $\times$   &   $\times$        \\
SeRum~\cite{cao2023attention}  & $\checkmark$        & E2E, w/o Loc.           &   $\times$    &   $\times$      \\ 
UniDet~\cite{long2022towards}  & $\checkmark$        & $\times$           &   $\times$    & $\checkmark$      \\ 
\midrule
\ourmodel                      & $\checkmark$        & E2E           & E2E (TSR + TCR) & $\checkmark$   \\ \bottomrule
\end{tabular}
\end{adjustbox}

\label{table:unified}
\end{table}

\section{Related works}
\label{sec:rela}

    \subsection{Scene Text Spotting} 
    Scene text spotting typically employs a unified end-to-end trainable network, blending text detection and text recognition into a cross-modal assisted paradigm. This integrated approach streamlines text detection and recognition into a singular network. It enables simultaneous localization and identification of text within images, capitalizing on the synergistic relationship between text detection and recognition to augment overall performance. Scene text spotting can be broadly classified into two main categories: regular end-to-end scene text spotting and arbitrarily-shaped end-to-end scene text spotting. Regular end-to-end scene text spotting concentrates on detecting and recognizing text within rectangular or standard-shaped regions, whereas arbitrarily-shaped end-to-end scene text spotting broadens its scope to handle text in irregular or curved shapes.

    \noindent\textbf{Regular End-to-end Scene Text Spotting.}
    Li et al.~\cite{li2017towards} introduced one of the earliest end-to-end trainable scene text spotting methods, which effectively integrated box text detection and recognition features using RoI Pooling~\cite{ren2015faster} within a two-stage framework. Originally designed for horizontal and focused text, their method showed significant performance improvements in an enhanced version~\cite{li2019towards}. Busta et al.~\cite{busta2017deep} also contributed to this area with their end-to-end deep text spotter, which further advanced the integration of detection and recognition. In subsequent developments, He et al.\cite{he2018end} and Liu et al.\cite{liu2018fots} incorporated anchor-free mechanisms to enhance both the training and inference speed. They employed novel sampling strategies, such as RoI-Rotate, to extract features from quadrilateral detection results, further refining the end-to-end framework.

    \noindent\textbf{Arbitrarily-shaped End-to-end Scene Text Spotting.} 
    Liao et al.~\cite{lyu2018mask} introduced Mask TextSpotter leveraging Mask R-CNN with character-level supervision to detect and recognize arbitrarily-shaped text. Mask TextSpotterv2~\cite{liao2019mask} reduced the dependence on character-level annotations, improving efficiency. Qin et al.~\cite{qin2019towards} employed RoI Masking to focus attention on arbitrarily-shaped text regions. Feng et al.~\cite{feng2019textdragon} utilized RoISlide for handling long text, whereas Wang et al.~\cite{wang2020all} focused on boundary points detection, text rectification, and recognition. CharNet~\cite{xing2019convolutional} also catered to arbitrarily-shaped text spotting. Segmentation Proposal Network (SPN)~\cite{Liao2020MaskTV} and ABCNet~\cite{Liu2020ABCNetRS} are other noteworthy contributions. ABINet++~\cite{Fang2022ABINetAB} innovatively used a vision model and a language model with an iterative correction mechanism. SwinTextSpotter~\cite{huang2022swintextspotter} used a Transformer encoder for detection and recognition. Approaches based on DETR~\cite{Carion2020EndtoEndOD} and variants~\cite{Zhu2020DeformableDD} for RoI-free scene text spotting have also shown promising results. TESTR~\cite{Zhang2022TextST} used an encoder and dual decoders, while TTS~\cite{Kittenplon2022TowardsWT} used a transformer-based approach. SPTS~\cite{Peng2021SPTSST} and variants~\cite{liu2023spts} employed a single point for each instance and used a Transformer to predict sequences. DeepSolo~\cite{ye2023deepsolo} allows a single decoder to perform text detection and recognition. ESTextSpotter~\cite{Huang2023ESTextSpotterTB} introduced an explicit synergy-based Transformer model with task-aware queries and a vision-language communication module to enhance scene text spotting. oCLIP~\cite{Xue2022LanguageMA} boosts text spotting performance via predefined pretext contrastive learning tasks. TCM~\cite{Yu2023TurningAC} and variants~\cite{Yu2024TurningAC} designed a flexible framework to turn a CLIP~\cite{Radford2021LearningTV} model into a text detector and spotter. BridgeSpotter~\cite{Huang_2024_CVPR} connects a fixed detector and recognizer to retain modularity while improving end-to-end optimization for text spotting. DNTextSpotter~\cite{DNTextSpotter} introduced a novel denoising training method that aligns text content and position using Bezier curve-based positional queries. WeCromCL~\cite{Wu2024WeCromCLWS} proposed a weakly supervised cross-modality contrastive learning framework that detects transcriptions without location annotations by modeling character-wise appearance consistency. InstructOCR~\cite{Duan2024InstructOCRIB} leveraged human language instructions to enhance scene text spotting, improving text interpretation and performance on OCR-related tasks. 

    \subsection{Key Information Extraction} 
Existing KIE approaches can be roughly categorized into two types: OCR-dependent models and OCR-free models. OCR-dependent models focus on using OCR for extracting textual information. Early KIE methods primarily built layout-aware or graph-based representations for KIE tasks via sequence labeling with OCR inputs~\cite{liao2022real,shi2016end,da2023multi,xu2020layoutlm, xu2021layoutlmv2, huang2022layoutlmv3, xu2021layoutxlm,li2021structurallm,appalaraju2021docformer,li2021selfdoc,gu2021unidoc,gu2022xylayoutlm,lee2022formnet,peng2022ernie,luo2023geolayoutlm,Yu2020PICKPK}. However, these methods often rely on text with a proper reading order or the use of extra modules~\cite{wang2021layoutreader, zhang2023reading} for OCR serialization, which is not always practical in real-world scenarios where the layout may be complex or unordered.
To address the serialization issue, some methods leverage additional detection or linking modules to model the complex relationships between text blocks or tokens~\cite{hwang2021spatial,xu2021layoutxlm,hong2022bros,yu2022structextv2,luo2023geolayoutlm,yang2023modeling,zhang2023reading,wei2023ppn,Yu2023ICDAR2C}. While these strategies mitigate the reading order problem, the added complexity of decoding and post-processing steps often limits their generalization ability, making them less adaptable to a wide variety of document layouts.
In contrast, generation-based methods~\cite{tang2023unifying,cao2023genkie,cao2022query} are proposed to alleviate the burden of post-processing and task-specific link designs. Another category of OCR-free methods offers an alternative by either utilizing OCR-aware pre-training or by incorporating OCR modules within an end-to-end framework.
For example, Pix2Struct~\cite{Lee2022Pix2StructSP} proposed a pretraining task in which the model generates a complete HTML DOM tree based on a masked webpage screenshot, without relying on OCR.
PreSTU~\cite{Kil2022PreSTUPF} introduced an OCR-aware pre-training objective that directly generates text sequences from image-based inputs. Donut and other Seq2Seq-like methods~\cite{kim2022donut,davis2022end,dhouib2023docparser,cao2023attention} adopted a text reading pre-training objective and generate structured outputs consisting of text and entity tokens.
By explicitly equipping text reading modules, previous work~\cite{wang2021towards,tang2021matchvie,zhang2020trie,kuang2023visual,yu2022structextv2} can maintain high performance in end-to-end KIE with task-specific design.

\subsection{Table Recognition}
Tables, as structured data, provide a succinct and compact format for organizing valuable content. Recent advancements in vision-based approaches have significantly improved the extraction of tables from documents. To provide a comprehensive overview, the task of table extraction from documents is typically divided into three main processes: table detection, table structure recognition, and table content recognition. Table detection, primarily concerned with locating tables within documents or images, has been extensively explored in previous work~\cite{staar2018corpus, zhong2019publaynet}, though it is beyond the scope of this paper. Table structure recognition (TSR) involves identifying the structure of a given table within a document or images, and has long been a focal point in the document understanding community~\cite{gobel2013icdar,gao2019icdar,kayal2021icdar,Raja2020TableSR,liu2021show,GTE}. Table content recognition (TCR) focuses on locating and recognizing text instances within the table cells and can be accomplished using established offline OCR models. In this paper, we concentrate on table recognition (TR) tasks that integrate both table structure recognition and table content recognition. Table recognition methods can be broadly categorized into two groups: non-end-to-end-based~\cite{TableMaster,tableformer,Tsrformer,TRUST,gridformer,VAST} and end-to-end-based~\cite{EDD,ly2023end} approaches. Non-end-to-end-based methods mainly first recover table structure via a specific model and then employ offline OCR models to construct complete HTML sequences via complex post-process. It is worth noting that end-to-end-based table recognition tasks remain less explored due to their complexity and challenging nature. Benefiting from the modularized architecture design, our model effectively separates the extraction of pure table HTML tags with cell text center point sequences and the cell text recognition sequences, accomplishing table recognition in an end-to-end fashion.
    
\subsection{Layout Analysis}
Geometric layout analysis focuses on detecting semantically coherent text blocks as objects~\cite{zhong2019publaynet,long2022towards,hts}. Recent approaches have modeled this task using various techniques, including object detection~\cite{Schreiber2017DeepDeSRTDL}, semantic segmentation~\cite{long2022towards}, and graph-based learning over OCR token structures via Graph Convolutional Networks (GCN). For example, Unified Detector~\cite{long2022towards} utilized segmentation-based formulations to pursue unifying scene text detection and layout analysis through an affinity matrix for modeling grouping relations, but it cannot generate word-level entities and lacks recognition capabilities. Another direction in layout analysis focuses on semantic parsing of documents to extract key-value pairs~\cite{li2021structurallm,long2022towards}. These methods typically leverage language models built on top of OCR outputs. In this paper, we conduct geometric layout analysis for identifying a 3-level hierarchical structure: words, lines, and paragraphs in an end-to-end unified paradigm.

\subsection{Unified Frameworks}
There is an increasing shift towards developing unified frameworks for parsing text-rich images across multiple tasks. Earlier works, such as DocReL~\cite{li2022relational} and BROS~\cite{hong2022bros} model relations between table cells or entities through binary classification or a relational matrix, which also requires an off-the-shelf OCR engine.
StrucTexTv2~\cite{yu2022structextv2} proposed a multi-modal learning framework aimed at document image understanding tasks by constructing self-supervised tasks. Yet, it requires several task-specific designs for downstream tasks, like Cascade R-CNN for table cell detection. Additionally, SeRum~\cite{cao2023attention} converts the end-to-end KIE task into a local decoding process and then shows its effectiveness on text spotting task. SCOB~\cite{kim2023scob} achieves universal text understanding across tasks by using character-wise supervised contrastive learning with online text rendering, effectively bridging domain gaps in document and scene text images with weak supervision. DocRes\cite{Zhang2024DocResAG} introduced a dynamic, task-specific prompt to unify five document image restoration tasks, while UPOCR\cite{Peng2023UPOCRTU} unifies various OCR pixel-level tasks within a single image-to-image transformation framework. Recent efforts have focused on developing more general, unified parsing frameworks. StrucTexTv3~\cite{Lyu2024StrucTexTv3AE}, integrated a multi-scale visual transformer, a multi-granularity token sampler, and instruction learning, achieving state-of-the-art results in text perception and comprehension tasks. DocOwl1.5~\cite{Hu2024mPLUGDocOwl1U} introduces a unified structure learning framework with H-Reducer to enhance MLLMs for document understanding. UDOP~\cite{tang2023unifying} integrates visual, textual, and layout modalities into a universal architecture for document processing. Nougat~\cite{Blecher2023NougatNO} proposes a Transformer-based OCR model specialized for academic documents. DocLayLLM~\cite{Liao2024DocLayLLMAE} presents a lightweight MLLMs for document layout understanding. KOSMOS 2.5~\cite{Lv2023Kosmos25AM} enables document OCR by aligning vision and language modalities through caption grounding and region-text associations. GOT~\cite{wei2024general} proposed a unified end-to-end OCR model capable of processing diverse optical signals (e.g., text, formulas, tables) across multiple tasks. It leveraged a high-compression encoder and long-context decoder for handling both scene and document images effectively. However, it lacks text localization ability. We note that there is a contemporaneous work with the same name, Microsoft's \emph{OmniParser V2}\footnote{https://github.com/microsoft/OmniParser}, which focuses on GUI grounding and computer-use agents, while our work studies unified VsTP for text-rich image understanding.

In this paper, we introduce \ourmodel, a unified framework designed to perform a wide range of visually-situated parsing tasks in an end-to-end manner. These tasks include text spotting, key information extraction, table recognition, and layout analysis, all of which are consolidated within a unified framework. 
\ourmodel can represent the heterogeneous structures of text in natural scenes or document images by decoupling structured points with text regions and contents. 
This two-stage approach caters to the intrinsic characteristics of text-rich images where the text instances can be parsed concurrently, thereby facilitating an enhancement in universality. Transformer-based models like OmniParser V2 incur high computational and spatial costs, limiting edge deployment. Recent works on distributed inference and model partitioning offer orthogonal solutions \cite{Yao2025EnergyEfficientEI, Li2024DistributedDI, Xu2022EdgeLF}, which will be explored in future work.

\subsection{Comparison to the Conference Version}
This paper presents a substantial extension of our previous work~\cite{omniparser}, incorporating three key advancements that contribute to the advancement in the area of unified models for visual text parsing.

\begin{enumerate}

\item \ourmodel addresses the limitations of our previous conference version, which used three separate decoders for structure point sequence generation, detection, and recognition. These decoders, though sharing the same architecture, had independent parameters, increasing model size and computational costs. In contrast, OMNIPARSER V2 employs a token-router-based shared decoder with a simplified MoE transformer, reducing model complexity. This shared decoder improves efficiency and synergy across tasks, resulting in a 23.6\% reduction in model size and enhanced performance.

\item Our method exhibits strong adaptability across diverse tasks. In addition to text spotting, key information extraction, and table recognition, which were evaluated in the conference version, \ourmodel further validates the scalability and effectiveness of structured-points-of-thought prompting through detailed experiments on layout analysis tasks using the HierText dataset, reinforcing its capability to handle complex document structures.

\item We further investigate the incorporation of SPOT prompting into multimodal large language models (MLLMs) for VsTP, a domain where MLLMs traditionally underperformed~\cite{Liu2023OCRBenchOT,Yang2024CCOCRAC,fu2024ocrbenchv2}. Our experiments reveal substantial improvements in visual text parsing, highlighting the broad applicability and effectiveness of SPOT prompting across different model architectures.

\end{enumerate}

\begin{figure*}[htbp]
    \centering
    \centerline{\includegraphics[width=1.0\linewidth]{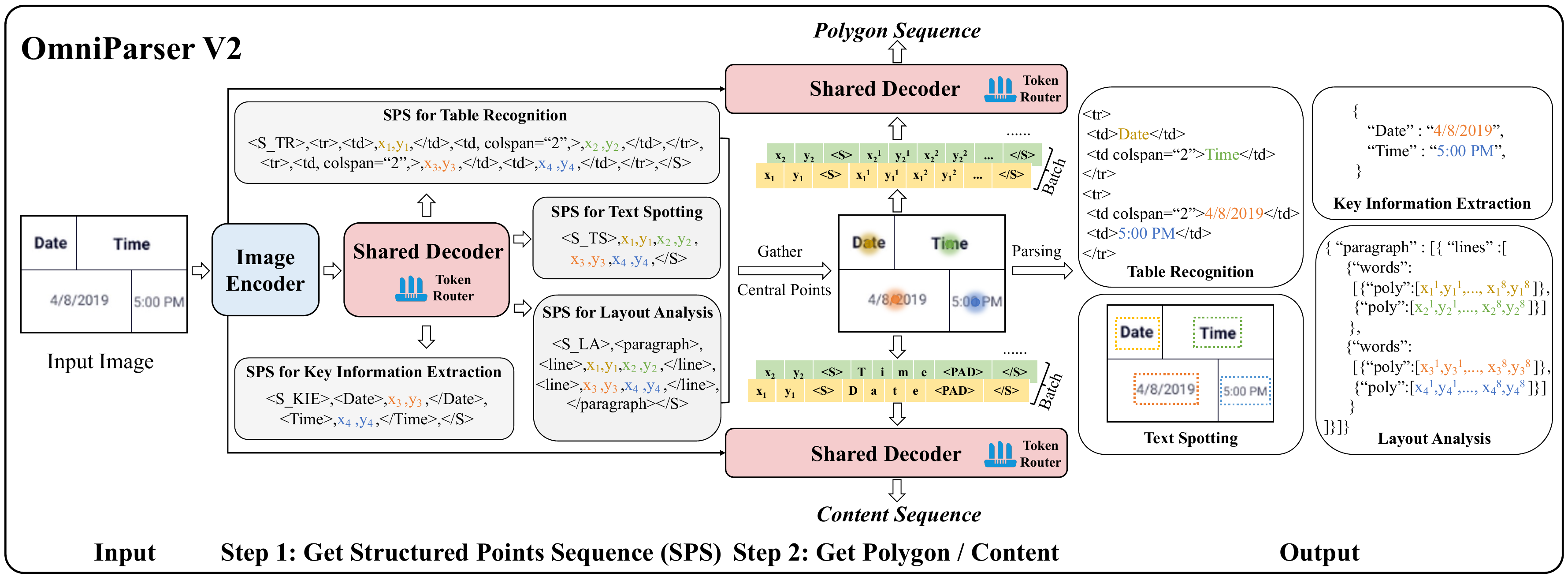}}
    \captionsetup{width=1.0\textwidth}
    \caption{\textbf{Schematic illustration of the proposed OmniParser V2 framework.} The token-router-based shared decoder homogenizes four tasks through a unified structural points representation without designing task-specific branches. 
    Furthermore, benefiting from decoupling points with content recognition and region prediction, the token-router-based shared decoder can generate polygonal contour and text content in parallel given the text points. SPS short for structured points sequence.}
    \label{fig:diagram}
\end{figure*}

\section{Methodology}
    \label{sec:method}

\subsection{Overview of OmniParsers V2}
\begin{figure}[htbp]
    \centering
    \includegraphics[width=0.45\textwidth]{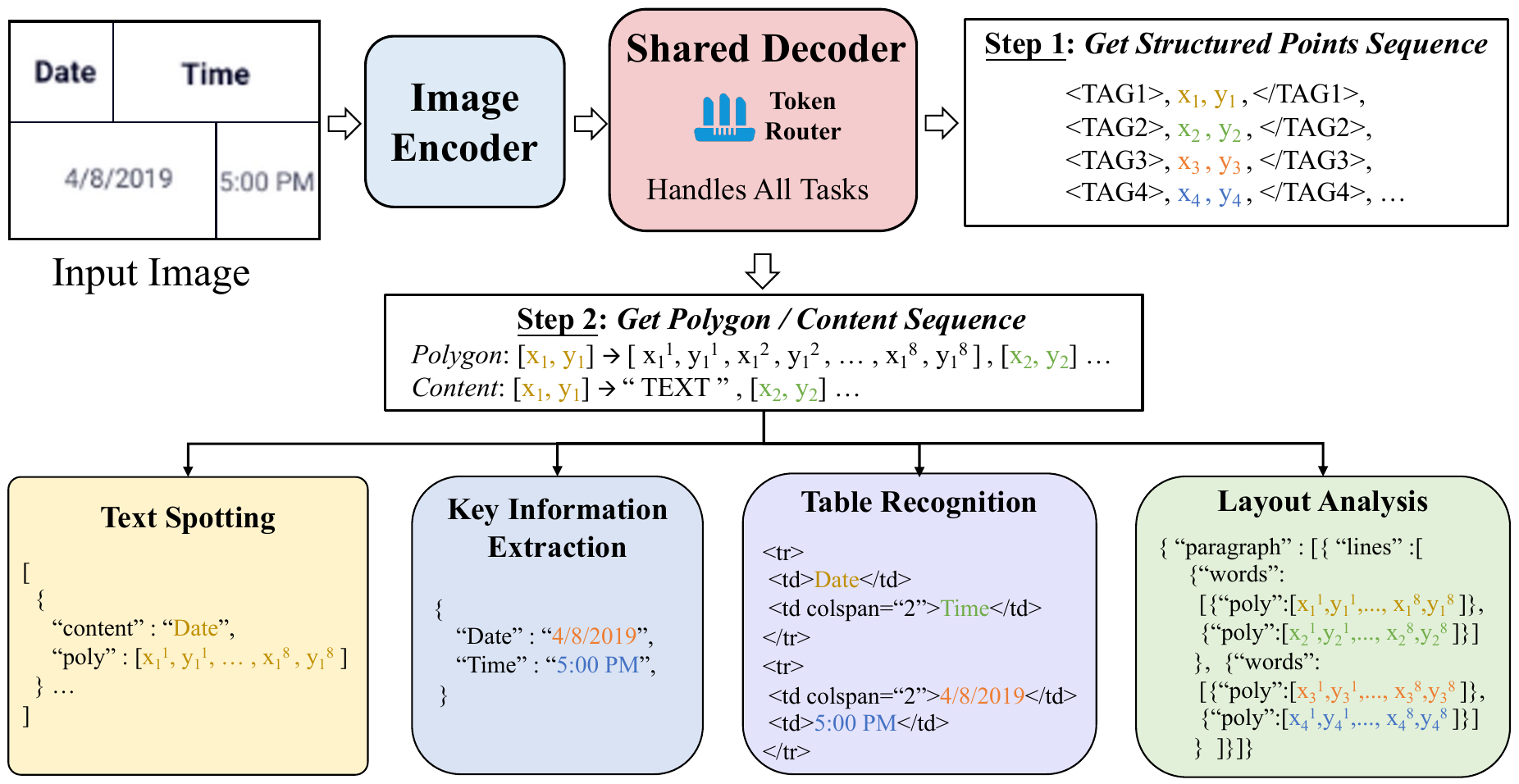}
    \caption{  \textbf{An overview of OmniParser V2’s two-stage pipeline.} Stage 1 extracts Structured Points Sequence (SPS) for each task, combining task-specific tags with text center points. Stage 2 uses these center points to decode either bounding polygons or text content. The unified shared decoder handles all tasks including text spotting, key information extraction, table recognition, and layout analysis. }
    \label{fig:explain_method}
\end{figure}

To enhance the clarity and intuitiveness of our methodology, we provide an illustrative overview of our proposed two-stage framework, OmniParser V2, as shown in ~\cref{fig:explain_method}. Our framework operates in two stages. First, the model generates a Structured Points Sequence (SPS), which serves as a unified representation combining visual layout (i.e., the center points of text instances) and task-specific structural semantics (e.g., tags like \texttt{<Date>}, \texttt{<td>}, or \texttt{<line>}). This SPS abstraction allows the model to represent various document parsing tasks in a unified and interpretable format. In the second stage, the model utilizes the same shared decoder to predict the final output for each point, either as a polygon (for localization) or textual content (for recognition), depending on the target task. To further clarify the construction logic of SPS across different tasks, we provide detailed task-specific SPS construction examples in ~\cref{fig:diagram} and in the supplementary material (Fig. 1 and Fig. 2).

\subsection{Task Unification}
As shown in~\cref{fig:diagram}, we propose a unified interface that represents structured sequences with three sub-sequences across diverse tasks.
Points are employed as bridges to effectively link structural tags with region and content sequences.

\mypara{Structured Points Sequence Construction}
comprises center points tokens as well as a variety of structural tokens designed for different tasks.
The x and y coordinates of each point are first normalized to the width and height of the image, respectively. Subsequently, they are quantized into discrete tokens within the range of $[0, n_{bins} - 1]$.
Moreover, structural tokens are introduced to represent the entire sequence, such as \token{address} in KIE task, \token{tr} in table recognition task, and \token{line} in layout analysis task. The constructed structural tokens serve as abstract semantic markers and do not contain or encode any image content themselves.
Note that text spotting can be seen as a special case that no structural token is incorporated. The bin size $n_{bins}$ means the number of coordinate vocab, which is set to 1,000.

\mypara{Polygon \& Content Sequence Construction}
is consistent across all tasks.
We represent curved text instances, referring to individual occurrences of text regions in the image, using a 16-point polygonal format and horizontal text instances using a 4-point bounding box format.
Each point in the polygon sequence is tokenized following the same procedure as the center point tokenization.
Besides, the transcription of text instances is converted into discrete tokens through char-level tokenization.

\subsection{Unified Architecture}

In light of our overarching goal to enhance the general-purpose paradigm for parsing text-rich images, we utilize a straightforward framework to assess the effectiveness of our proposed representation. To this end, we propose an encoder-decoder architecture that effectively addresses a wide range of visual text parsing tasks, as depicted in~\cref{fig:diagram}.

\mypara{Image Encoder.}
We adopt the Swin-B~\cite{liu2021swin} pre-trained on ImageNet 22k dataset as the fundamental visual feature extractor. 
Specifically, given an image $\mathbf{I} \in \mathbb{R}^{H \times W \times 3}$, we first use the image encoder to extract block-wise visual features which have strides of {4, 8, 16, 32} with respect to the input image. 
Afterward, we employ FPN~\cite{lin2017feature} for feature fusion in order to better capture text features at various scales, following~\cite{song2022vision}. 
Formally, a set of visual embeddings $\left\{\mathbf{v}_i \mid \mathbf{v}_i \in \mathbb{R}^d, 1 \leq i \leq n\right\}$ is generated, where $n$ is feature map size after FPN and $d$ is the dimension of the latent embeddings of the decoders.

\begin{figure}[htbp]
    \centering
    \includegraphics[width=0.49\textwidth]{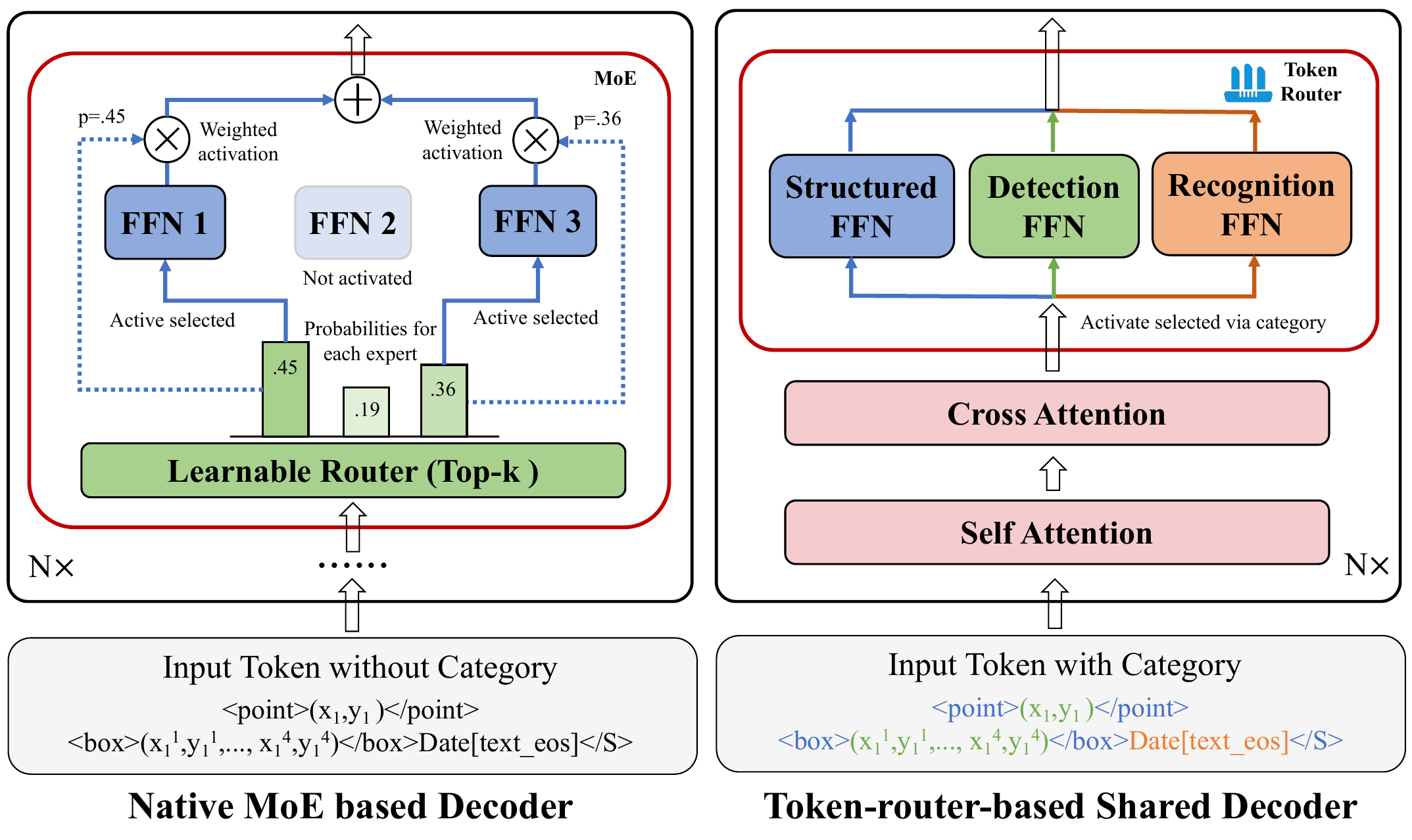}
    \caption{\textbf{Comparison of the token-router-based shared decoder and the native MoE based decoder.} Note that \textit{Self Attention} and \textit{Cross Attention} of the MoE based decoder, \textit{Add}, and \textit{LayerNorm} layers are omitted for easy visualization. The Structured FFN, Detection FFN, and Recognition FFN have separate parameters, while all other modules within the shared decoder utilize the same parameters. Input token from different categories is routed through their corresponding class-specific FFNs. In the right figure, different colors are different token categories and indicate the mapping between the input token and their respective FFNs.} 
    \label{fig:method_token_router}
\end{figure}

\mypara{Shared Decoders.}
The \pointsdecoder is used for structured points sequence generation, detection, and recognition, respectively. As shown in \cref{fig:method_token_router}, the shared decoder includes four Transformer decoder layers, each with eight heads and pre-attention layer normalization~\cite{xiong2020layer}. 
Each decoder layer integrates a token-router module with three category-specific FFNs (structured, detection, and recognition), forming a lightweight and deterministic alternative to traditional mixture-of-experts (MoE)~\cite{Shazeer2017OutrageouslyLN,NEURIPS2022_d46662aa,kil2023towards}.
Unlike conventional MoE designs, which rely on soft routing via learned probabilistic scores, our decoder employs task-aware token routing based on explicit category priors embedded within the input sequence. Each token $\mathbf{\tilde{s}}_j$ is associated with a predefined category (structured, detection, or recognition), which directly activates the corresponding FFN branch in the decoder without gradient-based router training. This deterministic routing scheme simplifies training, eliminates expert balancing overhead, and enhances interpretability of token-specialist assignments in multi-task settings.
The hidden dimension of each decoder layer and amplification factor for the MLP layer are set to 512 and 4, respectively. 
Due to varying maximum decoding lengths for the shared decoder, we assign uniquely randomly initialized positional encodings to the shared decoder, aiming to better model the dependencies within the sequences.

\mypara{Objective.}
During pre-training and fine-tuning, the model is trained by minimizing negative log-likelihood given the input sequence $\mathbf{s}$ and visual embeddings $\mathbf{v}$ at $j^{\text{th}}$ time step,

\begin{equation}
L=-\sum_{j=k}^N w_j \log P\left(\mathbf{\tilde{s}}_j \mid \mathbf{v}, \mathbf{s}_{k: j-1}\right) \,,
\label{eq:loss}
\end{equation}
where $\mathbf{\tilde{s}}$ denote the target sequence and $N$ is the length of the sequence. 
Additionally, $w_j$ is the weight value for the $j^{\text{th}}$ token.
We empirically set $w$ to $4.0$ for structural or entity tags and $1.0$ for other tokens.
First $k$ prompt tokens are excluded from the loss calculation.

\subsection{Pre-training Methods}

In our framework, generating structural points sequence is more challenging as it requires the \pointsdecoder to understand the text structure and reason entity semantics with image-based input only. 
Therefore, we adopt spatial-aware and content-aware pre-training strategies: spatial-window prompting and prefix-window prompting, to enhance richer spatial and semantic representation learning. 

\begin{figure}[htbp]
    \centering 
    \centerline{\includegraphics[width=1.0\linewidth]{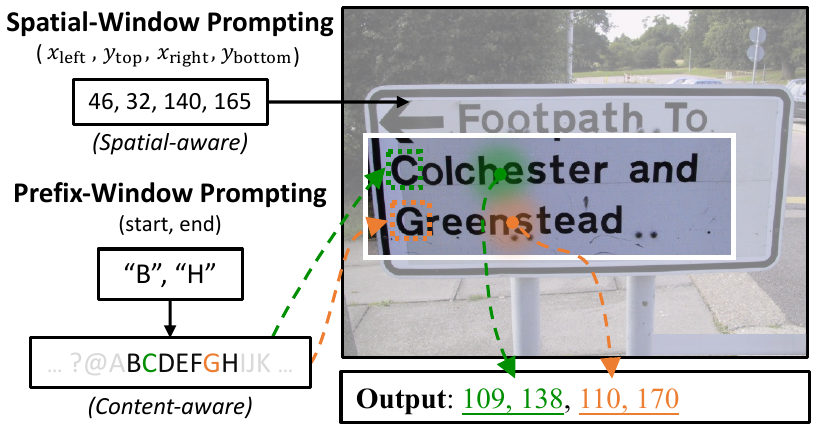}}
    \caption{\textbf{Spatial-Window Prompting} utilizes a 2-point prompt, denoted as $(x_{\texttt{left}}, y_{\texttt{top}}, x_{\texttt{right}}, y_{\texttt{bottom}})$, to specify the location of the prompting spatial window. \textbf{Prefix-Window Prompting} employs a 2-character prompt indicating the starting and ending characters of the prefix-window within the entire dictionary. The selected prefix range is highlighted in \textbf{black}, while others are shaded in \textcolor{gray}{gray}. The outputs are the center points of two words: \text{``Colchester''} and \text{``Greenstead''}, as their corresponding prefixes alphabets, \text{``C''} and \text{``G''} fall within the predefined prefix alphabet range [\text{``B''}, \text{``H''}], but the word \text{``and''}  is excluded because its prefix alphabet falls outside this range.}
    \label{fig:prompting}
\end{figure}

\mypara{Spatial-Window Prompting} guides the \pointsdecoder to read text inside a specified window.
As shown in~\cref{fig:prompting}, only the text center point located in the specified window is considered during training. 
The spatial-window prompting mechanism consists of two patterns: fixed pattern and random pattern.
In the fixed pattern, the window is uniformly sampled from a list of pre-defined layouts, such as $3\times3$ or $2\times2$ grids.
In the random pattern, the window is randomly sampled from an image, ensuring it covers at least $1/9$ of the image. 
More details are provided in Section 2.1 of the supplementary material.
Similar to Starting-Point Prompting~\cite{kil2023towards}, this spatial-aware prompting strategy allows the detection of numerous text from images, even with a limited decoder length.

\mypara{Prefix-Window Prompting} guides the \pointsdecoder to output center points of text with a specified single char prefix.
This strategy aims to instruct the model in locating text instances whose single-character prefix falls within the designated prefix-window charset, while disregarding instances with prefixes outside this charset. The prefix-window charset is sampled from an ordered list of character dictionaries, including 26 uppercase letters, 26 non-capital lowercase, 10 digits, and 34 ASCII punctuation marks, defined by the starting and ending characters.
With the aid of prefix-window prompting, the \pointsdecoder can encode character-level semantics and thus achieve better performance for predicting complex text structures from various tasks such as KIE.

\begin{figure*}[htbp]
    \centering
    \centerline{\includegraphics[width=0.95\linewidth]{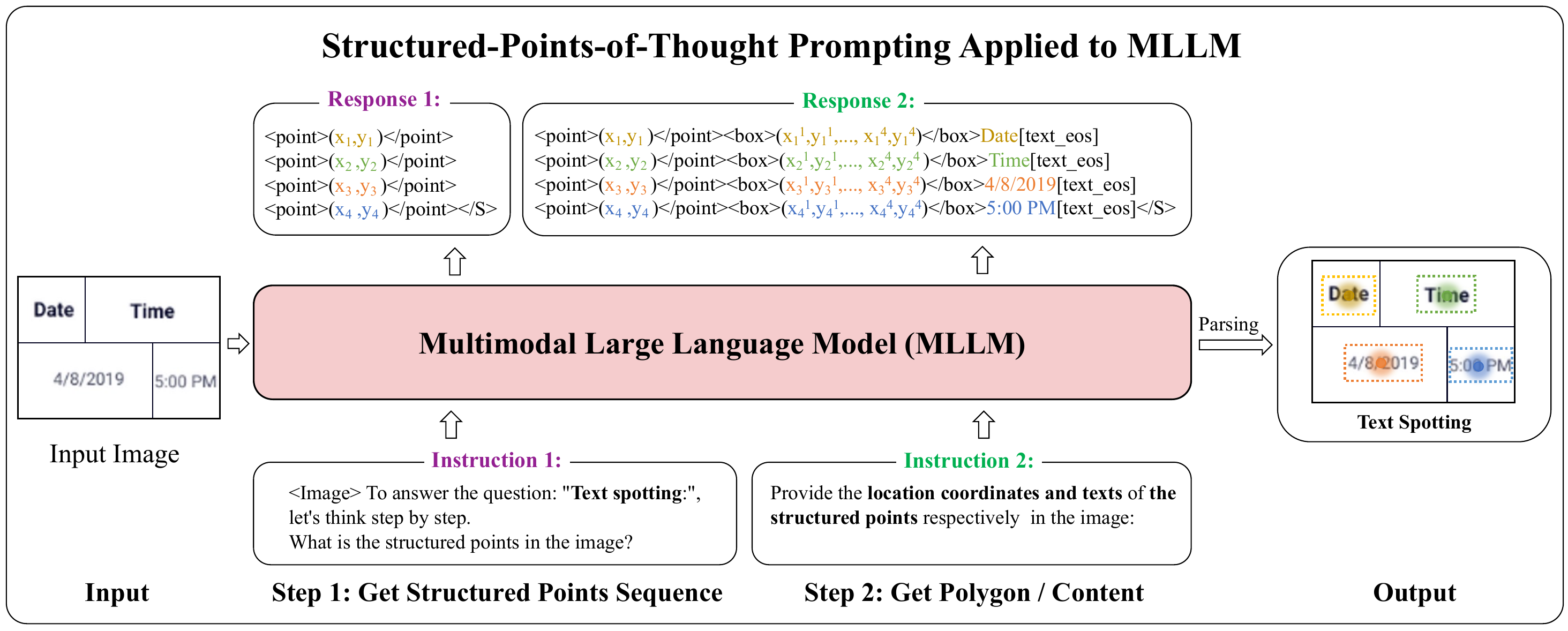}}
    \captionsetup{width=1.0\textwidth}
    \caption{\textbf{Illustration of the proposed structured-points-of-thought prompting applied to an existing multimodal large language model (MLLM) pipeline.} In the first conversation, the original image is combined with \textbf{Instruction 1} as a prompt, guiding the MLLM to generate a structured points sequence (\textbf{Response 1}), which represents the center points of each text instance in reading order for the text spotting task. In the second conversation, the first conversation is used as context, and \textbf{Instruction 2} prompts the MLLM to generate both the location coordinates and text content corresponding to each center point within the structured points sequence (\textbf{Response 2}). Finally, the extracted information is formatted to produce the expected text spotting results. In this figure, we use the text spotting task as an example, where the sequences corresponding to \textbf{Instruction 1} and \textbf{Response 1} are shown. In practice, when adapting the pipeline to the other three tasks, the term ``Text spotting'' in \textbf{Instruction 1} is replaced with the respective task name (key information extraction, table recognition, or layout analysis). Likewise, for these tasks, \textbf{Response 1} corresponds to the task-specific structured points  sequence (SPS).}
    \label{fig:mllm_diagram}
\end{figure*}

\subsection{SPOT Applied to MLLM}
We begin by providing a high-level overview of the motivation behind and the rationale for applying SPOT to Multimodal Large Language Models (MLLMs). Following this, we offer a comprehensive description of the implementation pipeline, curated dataset, fine-tuning process, and inference procedure for MLLMs. Note that the primary motivation for applying SPOT to MLLMs is not to directly replace domain-specific models like OmniParser V2, but rather to validate the generality and flexibility of SPOT prompting across architectures of different model scales. 

\mypara{Overview.}
Recently, chain-of-thought (CoT) prompting~\cite{Wei2022ChainOT} has emerged as an effective technique to improve the accuracy of model outputs. CoT prompting guides models through a structured reasoning process via multi-turn conversations, encouraging models to generate intermediate reasoning steps before providing the final answer. This approach has been particularly successful for more complex tasks, leading to improved accuracy. Our two-stage SPOT prompting can also be viewed as a form of CoT prompting. In the first stage, the model is prompted to generate a structured points sequence that identifies the center points of each text instance. In the second stage, the model generates the corresponding polygonal contours and content sequences separately.

\mypara{Pipeline.}
Intuitively, we propose the application of SPOT prompting to existing multimodal large language models (MLLMs) to address their limitations in visual text parsing tasks, such as text localization and recognition~\cite{Liu2023OCRBenchOT,Yang2024CCOCRAC,fu2024ocrbenchv2}. As illustrated in~\cref{fig:mllm_diagram}, we introduce an efficient, structured, and novel pipeline that enhances an MLLM’s ability to perform text spotting. Specifically, we instruct the model to first identify the center points of each text instance and generate a structured points sequence, followed by extracting the location coordinates and text content of each identified point. Finally, we can parse the last response to get each text instance's location and transcript. In this way, SPOT ensures the model consistently anchors its inferences to the center point of the text in the image, making the inference more grounded and traceable, thereby alleviating the hallucination problem of visual text parsing in MLLMs.

\begin{figure*}[t]
    \centering
    \centerline{\includegraphics[width=0.95\linewidth]{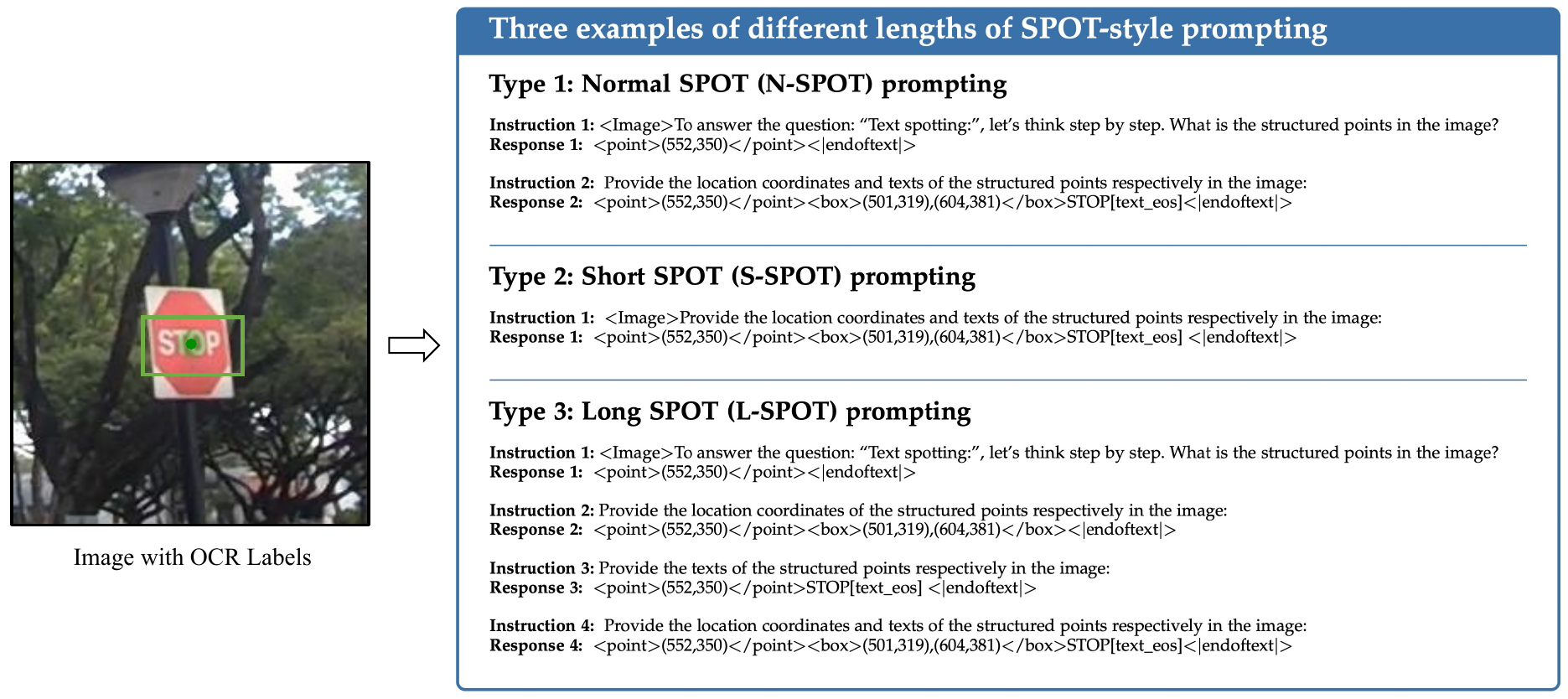}}
    \captionsetup{width=1.0\textwidth}
    \caption{\textbf{Examples of constructing SPOT-style prompting for supervised fine-tuning on MLLMs.} The figure demonstrates the construction of SPOT-style prompts using the QwenVL~\cite{Qwen-VL} format as an example for the text spotting task. Given an input image and the corresponding OCR labels, we generate three different lengths of SPOT prompting, including normal SPOT (N-SPOT), short SPOT (S-SPOT), and long SPOT (L-SPOT) prompting. Short SPOT prompting omits the intermediate generation of the structured points sequence, whereas long SPOT prompting includes additional steps, such as detection and recognition prompting. These prompts are subsequently used to fine-tune the MLLM.}
    \label{fig:spot_style_prompt_vis}
\end{figure*}

\mypara{Dataset.}
To instantiate this methodology, we curated a SPOT-style instruct-tuning dataset and performed supervised fine-tuning (SFT) on existing MLLMs to evaluate their performance in text spotting. Examples of the SPOT-style instruct-tuning data are provided in~\cref{fig:spot_style_prompt_vis}. Specifically, we performed ablation studies with three different lengths of SPOT prompting, referred to as normal SPOT (N-SPOT), short SPOT (S-SPOT), and long SPOT (L-SPOT) prompting, to fine-tune the MLLM. We used publicly available datasets and data collected from Platypus~\cite{Wang2024PlatypusAG} to construct the SPOT-style SFT datasets. The settings and sizes of our curated SFT data are shown in~\cref{tab:spot_like_sft_data}. For further information about the sources of each dataset subset within the SPOT-style SFT data, please refer to Section 1.5 of the supplementary material.

\begin{table}[htbp]
\centering
\caption{\textbf{The settings and number of our curated SFT data.} Num. short for number. }
   \begin{adjustbox}{max width=0.49\textwidth}
   \begin{threeparttable}
\begin{tabular}{lcccc}
\toprule
  Task Type                         & Prompt Setting     & Data Num.  & Data Name  \\
 \midrule
  \multirow{2}{*}{Text Spotting}  & \multirow{2}{*}{N-SPOT, S-SPOT, L-SPOT}            & 181,593    & TS180k \\
  &             & 389,433            & TS380k \\
   \midrule
  \multirow{2}{*}{Read All Text}  & \multirow{2}{*}{Read all the text in the image.}            & 446,702        & R440k \\
  &             & 981,284             & R980k \\

\bottomrule
\end{tabular}
   \end{threeparttable}
   \end{adjustbox}

\label{tab:spot_like_sft_data}
\end{table}

\mypara{Fine-tuning and Inference Procedures.}
During supervised fine-tuning, we employ the official fine-tuning scripts and configuration of the respective MLLM methods with our prepared training data. After training for up to three epochs, we evaluate the finetuned MLLM model on public text spotting datasets, following the evaluation metrics used in TextMonkey~\cite{textmonkey}. During the inference phase, we first feed the model Instruction 1, prompting it to generate the structured points sequence. Then, Response 1 is used as context for the next step, where Instruction 2 prompts the MLLM to generate both the location coordinates and text content for each center point in the structured points sequence. Finally, we format the extracted information to produce the expected text spotting results. When extending this procedure to the other three tasks, we similarly fine-tune MLLM methods for one epoch using its official training scripts and settings. The training datasets for fine-tuning correspond exactly to those used for the respective tasks in OmniParser V2, and the inference-time test sets and evaluation protocols likewise follow the task-specific settings.

\begin{table*}[htpb]
\centering
\caption{\textbf{Comparisons on text spotting task.} `S', `W', and `G' refer to the spotting performance obtained by utilizing strong, weak, and generic lexicons, respectively. The end-to-end metrics are highlighted as they are the primary metrics for text spotting. Bold and underline denote the first and second performances, respectively. $^*$ indicates the use of open-source code on our dataset configuration. `PT` means pre-training process. `SPT` indicates that, during the pre-training stage, our method uses the same amount of pre-training data as the compared models for a fair comparison. 
}
\resizebox{\textwidth}{!}{
\begin{tabular}{lccccccccccccccccc}
\toprule
\multirow{3}{*}{Methods}             & \multicolumn{5}{c}{Total-Text}   & \multicolumn{5}{c}{CTW1500}      & \multicolumn{6}{c}{ICDAR 2015}    & \multirow{3}[4]{*}{FPS}       \\ \cmidrule(lr){2-6} \cmidrule(lr){7-11} \cmidrule(lr){12-17}
& \multicolumn{3}{c}{Detection}   & \multicolumn{2}{c}{E2E}    & \multicolumn{3}{c}{Detection}   & \multicolumn{2}{c}{E2E} & \multicolumn{3}{c}{Detection}   & \multicolumn{3}{c}{E2E} &         \\
\cmidrule(lr){2-4} \cmidrule(lr){5-6} \cmidrule(lr){7-9}  \cmidrule(lr){10-11} \cmidrule(lr){12-14} \cmidrule(lr){15-17} 
                    & P    & R    & F    & None & Full & P    & R    & F    & None & Full & P    & R    & F    & S    & W    & G  &
                    \\ \midrule
TextDragon~\cite{feng2019textdragon}          & 85.6 & 75.7 & 80.3 & 48.8 & 74.8 & 82.8 & 84.5 & 83.6 & 39.7 & 72.4 & 92.5 & 83.8 & 87.9 & 82.5 & 78.3 & 65.2  & 2.6\\
CharNet~\cite{xing2019convolutional}             & 88.6 & 81.0 & 84.6 & 63.6 & -    & -    & -    & -    & -    & -    & 91.2 & 88.3 & 89.7 & 80.1 & 74.5 & 62.2 & 5.4 \\
TextPerceptron~\cite{qiao2020text}      & 88.8 & 81.8 & 85.2 & 69.7 & 78.3 & -    & -    & -    & 57.0 & -    & 92.3 & 82.5 & 87.1 & 80.5 & 76.6 & 65.1 & - \\
CRAFTS~\cite{baek2020character}              & 89.5 & 85.4 & 87.4 & 78.7 & -    & -    & -    & -    & -    & -    & 89.0 & 85.3 & 87.1 & 83.1 & 82.1 & 74.9 & 8.8\\
Boundary~\cite{wang2020all}            & 88.9 & 85.0 & 87.0 & 65.0 & 76.1 & -    & -    & -    & -    & -    & 89.8 & 87.5 & 88.6 & 79.7 & 75.2 & 64.1  & - \\
Mask TextSpotter v3~\cite{Liao2020MaskTV} & -    & -    & -    & 71.2 & 78.4 & -    & -    & -    & -    & -    & -    & -    & -    & 83.3 & 78.1 & 74.2 & 3.0\\
PGNet~\cite{wang2021pgnet}               & 85.5 & 86.8 & 86.1 & 63.1 & -    & -    & -    & -    & -    & -    & 91.8 & 84.8 & 88.2 & 83.3 & 78.3 & 63.5 & 38.2 \\
MANGO~\cite{qiao2021mango}               & -    & -    & -    & 72.9 & 83.6 & -    & -    & -    & 58.9 & 78.7 & -    & -    & -    & 85.4 & 80.1 & 73.9  & 8.4 \\
PAN++~\cite{wang2021pan++}               & -    & -    & -    & 68.6 & 78.6 & 87.1 & 81.0   & 84.0 & -    & -    & -    & -    & -    & 82.7 & 78.2 & 69.2 & 36.0 \\
ABCNet v2~\cite{Liu2021ABCNetVA}           & 90.2 & 84.1 & 87.0 & 70.4 & 78.1 & 83.8 & 85.6 & 84.7 & 57.5 & 77.2 & 90.4 & 86.0 & 88.1 & 82.7 & 78.5 & 73.0 & 10.0 \\
TPSNet~\cite{wang2022tpsnet}              & 90.2 & 86.8 & 88.5 & 76.1 & 82.3 & -    & -    & -    & 59.7 & 79.2 & -    & -    & -    & -    & -    & - & -   \\
ABINet++~\cite{Fang2022ABINetAB}            & -    & -    & -    & 77.6 & 84.5 & -    & -    & -    & 60.2 & 80.3 & -    & -    & -    & 84.1 & 80.4 & 75.4 & 10.6 \\ 
GLASS~\cite{ronen2022glass}               & 90.8 & 85.5 & 88.1 & 79.9 & 86.2 & -    & -    & -    & -    & -    & 86.9 & 84.5 & 85.7 & 84.7 & 80.1 & 76.3 & 2.7 \\
TESTR~\cite{Zhang2022TextST}               & 93.4 & 81.4 & 86.9 & 73.3 & 83.9 & 92.0 & 82.6 & 87.1 & 56.0 & 81.5 & 90.3 & 89.7 & 90.0 & 85.2 & 79.4 & 73.6 & 5.3 \\
SwinTextSpotter~\cite{huang2022swintextspotter}    & -    & -    & 88.0 & 74.3 & 84.1 & -    & -    & 88.0 & 51.8 & 77.0 & -    & -    & -    & 83.9 & 77.3 & 70.5 & 2.9 \\
SPTS~\cite{Peng2021SPTSST}                & -    & -    & -    & 74.2 & 82.4 & -    & -    & -    & 63.6 & 83.8 & -    & -    & -    & 77.5 & 70.2 & 65.8  & 0.6 \\
TTS~\cite{kittenplon2022towards}                 & -    & -    & -    & 78.2 & 86.3 & -    & -    & -    & -    & -    & -    & -    & -    & 85.2 & 81.7 & 77.4 & - \\
UNITS~\cite{kil2023towards}                 & -    & -    & 89.8    & 82.2 & 88.0 & -    & -    & 88.6    & 66.4    & 82.3    & 91.0    & 94.0    & 92.5    & 89.0 & 84.1 & 80.3 & 3.1 \\
DeepSolo~\cite{ye2023deepsolo}            & 93.2 & 84.6 & 88.7 & 82.5 & 88.7 & -    & -    & -    & 56.7 & -    & 92.5 & 87.2 & 89.8 & 88.0 & 83.5 & 79.1 & 17.0 \\
DeepSolo$^*$~\cite{ye2023deepsolo}            & 92.8  & 82.4  & 87.4 & 81.2 & 87.8 & 91.5    & 84.8   & 88.0    & 64.9 & 81.2    & 92.4 & 88.8 & 90.6 & 88.9 & \ranksecond{84.4}  & 79.5 & 17.0 \\
BridgeSpotter~\cite{Huang_2024_CVPR}            & 92.0  & 86.5  & 89.2 & 83.3 & 88.3 & 92.1    & 86.2   & 89.0    & \rankfirst{69.8} & 83.9    & 93.8 & 87.5 & 90.5 & 89.1 & 84.2  & \ranksecond{80.4} & 6.7 \\
OmniParser~\cite{omniparser}                & 88.4 & 88.6 & 88.5 & {84.0} & {88.9} & 87.9 & 87.6 & 87.8 & 66.8 & {85.1} & 90.3 & 91.0 & 90.7 & {89.6} & 84.5 & 79.9 & 2.6 \\
\midrule
 \ourmodel$^{PT}$                & 90.0 & 87.6 & 88.8 & {84.1} & {89.1} & 89.0 & 86.8 & 87.9 & {67.3} & {85.2} & 91.5 & 92.5 & 92.0 & {89.4} & {84.1} & \ranksecond{80.4} & 8.2 \\ 
\ourmodel$^{SPT}$                & 89.7 & 88.5 & 89.1 & \ranksecond{84.2} & \ranksecond{89.3} & 89.2 & 86.8 & 88.0 & {67.4} & \ranksecond{85.4} & 91.5 & 92.9 & 92.2 & \ranksecond{89.7} & {84.2} & \ranksecond{80.4}  & 8.2 \\
\ourmodel                & 90.6 & 88.2 & 89.4 & \rankfirst{84.3} & \rankfirst{89.5} & 89.7 & 87.1 & 88.4 & \ranksecond{67.9} & \rankfirst{85.5} & 91.7 & 92.9 & 92.3 & \rankfirst{89.9} & \rankfirst{84.5} & \rankfirst{80.6} & 8.2 \\
\bottomrule
\end{tabular}
}

\label{table:textspotting}
\end{table*}

\section{Experiments}
\label{sec:experiments}
In this section, we conduct both qualitative and quantitative experiments on standard benchmarks, to verify the effectiveness and advantages of the proposed \ourmodel.

\subsection{Implementation Details}
\mypara{Pre-training.}
\ourmodel is first trained on a hybrid dataset containing the training split of Curved SynthText~\cite{Liu2021ABCNetVA}, ICDAR 2013~\cite{Karatzas2013ICDAR2R}, ICDAR 2015~\cite{karatzas2015icdar}, MLT 2017~\cite{nayef2017icdar2017}, Total-Text~\cite{ch2020total}, TextOCR~\cite{singh2021textocr}, HierText~\cite{long2022towards}, COCO Text~\cite{gomez2017icdar2017}, and Open Image V5~\cite{krylov2021open}. 
To accelerate convergence, we adopt a two-stage pre-training strategy following Pix2seq~\cite{chen2021pix2seq}. In the first stage, the model is trained with a batch size of 128 and image resolution of $768\times768$ for 500k steps. Subsequently, we continue training for an additional 200k steps with a batch size of 16 and image resolution of $1920\times1920$. Both stages utilize the AdamW~\cite{loshchilov2018decoupled} optimizer, with initial learning rates of \num{5e-4} and \num{2.5e-4}, respectively. Warm-up schedule is used for the first 5k steps, after which the learning rate is linearly decayed to 0. For data augmentation, we employ instance-aware random cropping, random rotation between $-90^\circ$ and $90^\circ$, random resizing, and color jittering. 
During pre-training, the center points of text instances are arranged in a raster scan order.

\mypara{Fine-tuning.}
For all tasks, the model was fine-tuned on the corresponding training split dataset. For text spotting and KIE tasks, the model is fine-tuned for 20k and 200k steps, respectively, with a learning rate set to \num{1e-4}. 
For table recognition and layout analysis, the default maximum sequence lengths for structured points sequence and content sequence are set to 1,500 and 200, respectively. 
The model is trained up to 400k steps with the learning rate set to \num{1e-4}.
For all tasks, the cosine learning rate scheduler is utilized.
Besides, the spatial-window prompting and prefix-window prompting are modified as $[0, 0, n_{bins} - 1, n_{bins} - 1]$ and [${\texttt{char}_\texttt{first}}$, ${\texttt{char}_\texttt{last}}$] (`!' and `\texttt{\texttildelow}' in the dictionary) respectively, to cover full spatial and prefix range.

\mypara{Datasets and Evaluation Metrics.}     
Please refer to Section 1 of the supplementary material for comprehensive descriptions of the datasets and evaluation metrics used across all four tasks, including text spotting, KIE, table recognition, and layout analysis.

\subsection{Comparisons with State-of-The-Art}

\mypara{Text Spotting.}
In~\cref{table:textspotting}, we compare \ourmodel with previous text spotting approaches.
On arbitrarily shaped text datasets, Total-Text~\cite{ch2020total} and CTW1500~\cite{liu2019curved}, our method establishes new state-of-the-art under two end-to-end metrics.
Specifically, our method surpasses previous SOTA by +0.6\% and +0.4\% on Total-Text and CTW1500 respectively with lexicon-based evaluation, outperforming all the other competitors. On the multi-oriented text dataset ICDAR 2015, our method also outperforms previous approaches under the strong, weak, and generic lexicon settings. 
Notably, our approach achieves comparable detection results, while significantly surpassing previous work under end-to-end metrics, highlighting its robust text-spotting capabilities. We attribute this superior performance to the decoupling of the detection and recognition processes, which allows each sub-task to be optimized independently, thereby enhancing the overall end-to-end performance. In addition, we report the performance of the pre-trained model without any task-specific fine-tuning (OmniParser~V2$^{PT}$), which already surpasses several strong text-spotting baselines such as DeepSolo~\cite{ye2023deepsolo}, UNITS~\cite{kil2023towards}, and the original OmniParser on both Total-Text and CTW1500, further demonstrating the robustness and generalizability of our methods. To ensure fairness, we also present results from a controlled pre-training regime using the same data volume as other baselines (OmniParser~V2$^{SPT}$), which still achieves competitive performance across all benchmarks, confirming the effectiveness of our method beyond data scale. Besides, despite its two-stage nature, \ourmodel reaches 8.2 FPS, outperforming most existing Transformer-based models, e.g., UNITS~\cite{kil2023towards}, SPTS~\cite{Peng2021SPTSST}, thanks to a parallelizable second stage and efficient batching.

\begin{table}[t]
   \centering
      \caption{{\bf Comparisons of end-to-end methods on key information extraction.} `F1' denotes the field-level F1 score and `Acc' denotes the tree-edit-distance-based accuracy. $^\dagger$ Since the SROIE dataset does not provide the necessary point location for each entity word, we generate these locations for evaluation purposes.}
   \begin{adjustbox}{max width=0.46\textwidth}
   \begin{threeparttable}
     \centering
     \begin{tabular}{lcccccc}
     \toprule
     \multirow{2}{*}{Methods} & \multirowcell{2}{Localization \\ Ability} & \multicolumn{2}{c}{CORD} & \multicolumn{2}{c}{SROIE}  & \multirow{2}[2]{*}{FPS}  \\
      \cmidrule(lr){3-4} \cmidrule(lr){5-6}
      &  & F1 & Acc & F1 & Acc &   \\
      \midrule 
        TRIE~\cite{zhang2020trie}                     & $\checkmark$  & -    & -     & 82.1 & - & 1.7   \\
        SCOB~\cite{kim2023scob}                     & $\checkmark$  & -    & 88.5     & - & -  & 1.2 \\
        Donut~\cite{kim2022donut}                     & $\times$  & 84.1 & \rankfirst{90.9}  & 83.2  & {92.8} & 1.5 \\
        Dessurt~\cite{davis2022end}                   & $\times$   & 82.5 & -     & 84.9  & -  &-  \\
        DocParser~\cite{dhouib2023docparser}          & $\times$   & 84.5 & -     & \rankfirst{87.3} & -  & 1.8  \\
        SeRum~\cite{cao2023attention}                 & $\times$   & 80.5 & 85.8  & 85.6 & {92.8} & - \\
        OmniParser~\cite{omniparser}                 & $\checkmark$      & \ranksecond{84.8} & 88.0 & 85.6 & {93.6} & 2.8 \\
        \midrule
        \ourmodel$^{SPT}$                   & $\checkmark$      & \ranksecond{84.8} & {88.5} & {85.4}  & \ranksecond{93.8} & 3.1 \\
        \ourmodel                 & $\checkmark$      & \rankfirst{85.0} & \ranksecond{88.7} & \ranksecond{85.8}  & \rankfirst{94.0} & 3.1 \\
        
        \bottomrule
     \end{tabular}
   \end{threeparttable}
   \end{adjustbox}

   \label{table:kie}
\end{table}

\mypara{Key Information Extraction.}
~\cref{table:kie} reports the performance of KIE task, comparing our method to state-of-the-art end-to-end approaches on the CORD~\cite{park2019cord} and SROIE~\cite{huang2019icdar2019} datasets. 
We have exclusively reported SeRum\textsubscript{total}~\cite{cao2023attention} since all generation-based methods utilize a schema that encompasses the entire token sequence of all key information, making it directly comparable.
Our model achieves an $85.0\%$ field-level F1 score on CORD, outperforming previous generation-based methods. 
Additionally, our method achieves the best TED-based accuracy on SROIE, demonstrating superior performance in character-level prediction. Notably, the proposed paradigm ensures accurate localization, which is critical for detailed document analysis and correction, an area where other generation-based methods fall short.
Moreover, unlike previous studies that utilized a massive corpus of document data for pre-training, our model is pre-trained on scene text data only. 
This highlights the exceptional generalizability of our unified model. We further include a controlled pre-training variant, OmniParser~V2$^{\text{SPT}}$, trained with the same data scale as baselines. It achieves competitive performance, validating that our improvements stem from model design rather than data size. Our approach delivers an FPS of 3.1 on par with or faster than prior KIE methods, thanks to efficient parallelization in the second stage.

\begin{table}[htbp]
\centering
\caption{\textbf{Comparisons of end-to-end table recognition methods on PubTabNet~\cite{EDD} and FinTabNet~\cite{GTE} datasets.} * represents our reproduced results, where the model was finetuned on PubTabNet and FinTabNet, respectively.}
   \begin{adjustbox}{max width=0.47\textwidth}
   \begin{threeparttable}
\begin{tabular}{lcccccc}
\toprule
 \multicolumn{6}{c}{PubTabNet (PTN)}       \\
 \midrule
  Methods                         & Input Size      & Decoder Len.      & S-TEDS & TEDS & FPS \\
 \midrule
WYGIWYS~\cite{deng2017image}                                              & 512            & -                & -   & 78.6  & - \\
  Donut*~\cite{kim2022donut}                                              & 1,280            & 4,000                &  25.2  & 22.7  & 0.8 \\
  EDD~\cite{EDD}                                          & 512             & 1,800                & 89.9   & 88.3 & - \\
  OmniParser~\cite{omniparser}                                          & 1024             & 1,500                & 90.4   & 88.8  & 1.3 \\
 \midrule
       \multirow{1}{*}{ \ourmodel$^{SPT}$   }      & 1,024            & 1,500             & {90.3}  & {88.6} & 1.7 \\
       \multirow{1}{*}{\ourmodel}      & 1,024            & 1,500             & \textbf{90.5}  & \textbf{88.9} & 1.7 \\

\toprule

 \multicolumn{6}{c}{FinTabNet (FTN)}       \\
 
 \midrule
  Methods                            & Input Size      & Decoder Len.      & S-TEDS & TEDS & FPS  \\
 \midrule

  Donut*~\cite{kim2022donut}                                                  & 1,280            & 4,000                &  30.6  & 29.1  & 0.8 \\
  EDD~\cite{EDD}                                                    & 512             & 1,800                & 90.6   & -    &- \\
  OmniParser~\cite{omniparser}                                                    & 1024             & 1,500                & 91.5   & 89.7   & 1.3  \\
 \midrule
        \multirow{1}{*}{ \ourmodel$^{SPT}$   }      & 1,024            & 1,500             & {91.8}  & {89.9} & 1.7 \\
\multirow{1}{*}{\ourmodel}  & 1,024            & 1,500             & \textbf{93.2}  & \textbf{90.5} & 1.7 \\

\bottomrule
\end{tabular}
   \end{threeparttable}
   \end{adjustbox}

\label{tab:ptn_and_ftn}
\end{table}

\mypara{Table Recognition.}
In~\cref{tab:ptn_and_ftn}, we compare \ourmodel's performance with end-to-end table recognition models. We fine-tuned the OCR-free model Donut~\cite{kim2022donut} for table recognition using the official training configuration.
Experimental results show that \ourmodel consistently outperforms previous end-to-end methods in both TEDS and S-TEDS~\cite{EDD} on various datasets. 
Notably, non-end-to-end table structure recognition models~\cite{TableMaster, tableformer, Tsrformer, TRUST, gridformer, VAST} use bounding boxes of cell contents for model training and employ offline OCR models for constructing final complete HTML sequences. In contrast, \ourmodel utilizes points, achieving comparable results in an end-to-end manner, simplifying post-processing and requiring fewer annotations compared to box-based methods. Under the same pre-training scale as existing baselines, OmniParser V2$^{\text{SPT}}$ still delivers top-tier results on PTN and FTN, reaffirming that the gains are rooted in model design rather than data quantity. Moreover, our method achieves 1.7 FPS, exceeding existing Transformer-based models like Donut (0.8 FPS) and OmniParser (1.3 FPS), validating its practical efficiency.

\begin{table}[htbp]
   \centering
      \caption{{\bf Comparisons of geometric layout analysis methods on HierText dataset.} `PQ' denotes the panoptic quality. $^\dagger$stand for the results from~\cite{long2022towards}.}
  \setlength\tabcolsep{1.3pt}
   \begin{adjustbox}{max width=0.49\textwidth}
   \begin{threeparttable}
     \centering
     \begin{tabular}{lccccc}

     \toprule
     \multicolumn{6}{c}{HierText Validation Set}       \\
     \midrule
     \multirow{2}{*}{Methods} & \multirowcell{2}{End-to-End} & \multicolumn{1}{c}{Word-level} & \multicolumn{1}{c}{Line-level} & \multicolumn{1}{c}{Paragraph-level}  & \multirow{2}[2]{*}{FPS} \\
      \cmidrule(lr){3-3} \cmidrule(lr){4-4} \cmidrule(lr){5-5}
      &  & PQ & PQ & PQ & \\
      \midrule 
        UniDec~\cite{long2022towards}   & $\checkmark$  &   48.4  & 61.2 & 52.8  & 3.9 \\
        Hi-SAM-B~\cite{hisam}   & $\checkmark$          & 59.2   & 62.8 & 55.6  & 3.4  \\
        \midrule 
             \ourmodel$^{SPT}$                  & $\checkmark$       &   {59.5} & {63.0} & {55.7} & 4.7 \\
         \ourmodel                 & $\checkmark$       &   \textbf{60.0} & \textbf{63.4} & \textbf{55.9} & 4.7  \\

     \toprule
     \multicolumn{6}{c}{HierText Test Set}       \\
     \midrule
     \multirow{2}{*}{Methods} & \multirowcell{2}{End-to-End} & \multicolumn{1}{c}{Word-level} & \multicolumn{1}{c}{Line-level} & \multicolumn{1}{c}{Paragraph-level}  & \multirow{2}[2]{*}{FPS}  \\
      \cmidrule(lr){3-3} \cmidrule(lr){4-4} \cmidrule(lr){5-5}
      &  & PQ & PQ & PQ & \\
      \midrule 
        GCP API$^\dagger$   & unknown     &  - & 56.1 & 46.3 &-  \\
        GCN-PP$^\dagger$   & $\times$     & -  & 62.2 & 50.1 &-  \\
        Mask-RCNN-Cluster$^\dagger$   & $\times$     & -  & 62.2 & 51.6 &-  \\
        MaX-DeepLab-Cluster$^\dagger$   & $\times$     &  - & 62.2 & 52.5   &-\\
        UniDec~\cite{long2022towards}   & $\checkmark$      & 48.2  & 62.2 & 53.6 & 3.9  \\
        Hi-SAM-B~\cite{hisam}   & $\checkmark$      &  59.7 & 63.3 & 54.4   &3.4 \\
        \midrule 
         \ourmodel$^{SPT}$                  & $\checkmark$       &   {60.4} & {63.7} & {54.8} &4.7 \\
         \ourmodel                 & $\checkmark$       & \textbf{61.6}  & \textbf{64.5} & \textbf{55.2} &4.7 \\
        
        \bottomrule
     \end{tabular}
   \end{threeparttable}
   \end{adjustbox}

   \label{table:layout_analysis}
\end{table}

\mypara{Layout Analysis.}
We evaluated \ourmodel on the widely used HierText~\cite{long2022towards} dataset for layout analysis, focusing on word, line, and paragraph level grouping using the Panoptic Quality (PQ)~\cite{Kirillov2018PanopticS} metric. The results are shown in~\cref{table:layout_analysis}. 
On the validation set, \ourmodel surpasses Hi-SAM-B~\cite{hisam} by +0.8\%, +0.6\%, and +0.3\% PQ, in terms of word, line, and paragraph grouping, respectively. On the test set, it further outperforms all methods, with gains of +1.9\%, +1.2\%, and +0.8\% PQ over Hi-SAM-B, highlighting its superior capability in capturing hierarchical text structures.
Compared to non-end-to-end models like GCN-PP, Mask-RCNN-Cluster, and MaX-DeepLab-Cluster, which rely on offline OCR and bounding boxes, \ourmodel achieves higher scores with a fully end-to-end pipeline. To mitigate fairness concerns, we report OmniParser V2$^{\text{SPT}}$ trained with matched-scale pre-training data, which still performs competitively and preserves layout parsing gains, supporting the architectural advantage of our approach. In addition, our method runs at 4.7 FPS, outperforming UniDec (3.9 FPS) and Hi-SAM-B (3.4 FPS). 
These results demonstrate that \ourmodel, powered by SPOT prompting, effectively models geometric structures and hierarchical relationships, delivering state-of-the-art performance without relying on external OCR modules or complex bounding box annotations.

\subsection{Analysis}
In this section, we begin by conducting ablation experiments on crucial designs in \ourmodel. Besides, we provide visualizations on downstream tasks to illustrate the effectiveness of \ourmodel.

\begin{table}[htbp]
\centering
\caption{\textbf{Ablation of pre-training strategies} on text spotting. }
\begin{adjustbox}{max width=0.45\textwidth}
\begin{tabular}{ccccccc}
\toprule
\multicolumn{2}{c}{Window-Prompting}& \multicolumn{2}{c}{Total-Text} & \multicolumn{3}{c}{ICDAR 2015} \\
\cmidrule(lr){1-2} \cmidrule(lr){3-4} \cmidrule(lr){5-7}
Spatial-     & Prefix-    & None           & Full          & S        & W        & G   \\ \midrule
  &                  & 82.5  & 87.8  & 88.3 & 83.2  & 78.5          \\
  &     \checkmark  & 83.2  & 88.4  &88.5  &83.3          & 78.9         \\
  \checkmark        &  & 83.8  &88.9  & 89.4  & 84.3  & 79.8          \\ 
 \checkmark & \checkmark  & \textbf{84.3}  & \textbf{89.5} & \textbf{89.9}  & \textbf{84.5}    & \textbf{80.6}    \\
\bottomrule
\end{tabular}
\end{adjustbox}

\label{table:abpretraintask}
\end{table}

\begin{figure*}[htbp]
  \centering 
  \centerline{\includegraphics[width=1.0\linewidth]{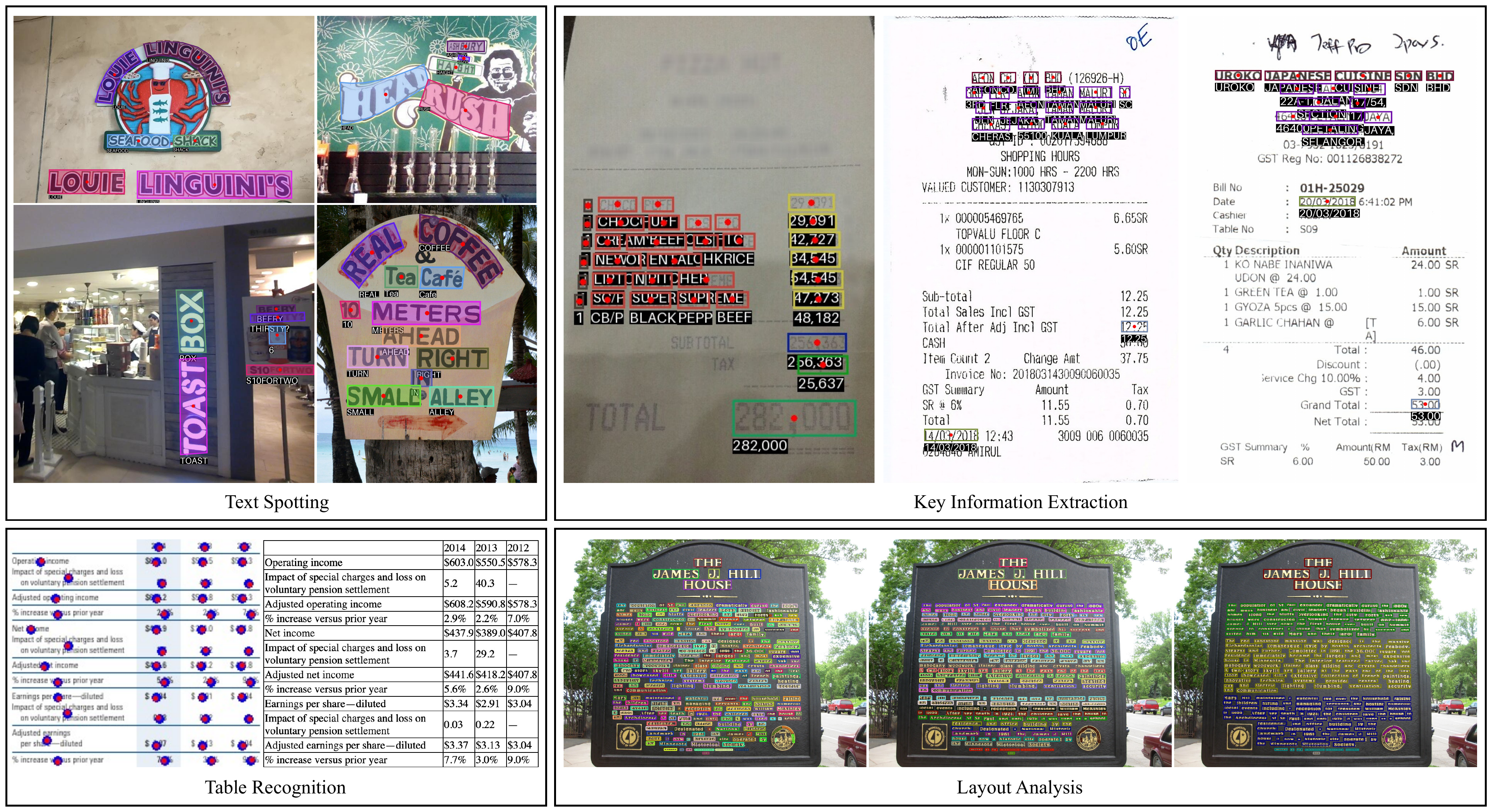}}
  \caption{ \textbf{Qualitative results} of text spotting, KIE, table recognition, and layout analysis. For KIE, points, polygons, and recognition are visualized. The color assigned to polygons indicates the entity type. For table recognition, we present point locations and a rendered table based on the prediction sequence, with an additional border for readability. Blue points and red points denote the GT and predicted points, respectively. For layout analysis, we show the detection results for word, line, and paragraph levels, respectively. Text instances belonging to the same hierarchical level are enclosed within rectangles of the same color. (The figure is best viewed in color.)}
  \label{fig:visualization}
\end{figure*}

\mypara{Ablating Pre-training Strategies.}
To examine the impact of spatial-window prompting and prefix-window prompting, we conducted ablative experiments, as shown in \cref{table:abpretraintask}.
Incorporating spatial-window prompting significantly improves model performance by enhancing the perception of spatial coordinate positions, resulting in more accurate predictions of structured points sequences. Similarly, adding prefix-window prompting boosts performance by enabling the model to capture diverse textual content within images, thereby enhancing its semantic understanding.
The spatial-window prompting and prefix-window prompting enhance the model's perception ability in coordinate space and semantic space, respectively. When both techniques are applied together, the model achieves state-of-the-art performance on both datasets, demonstrating their complementary effects in enhancing both coordinate and semantic space perception.

\begin{table}[htbp]
\centering
\caption{\textbf{Ablation of architectural designs and computational complexity} on the text spotting task. `BSL' represents the baseline method.}
\setlength\tabcolsep{1.3pt}
\begin{adjustbox}{max width=0.49\textwidth}
\begin{tabular}{ccccccccccc}
\toprule
\multirow{2}{*}{Method} & \multirowcell{2}{Visual \\ Backbone} & \multirow{2}{*}{Decoder} & \multicolumn{2}{c}{Total-Text} & \multicolumn{3}{c}{ICDAR 2015} &  \multirowcell{2}{ {Params} \\ {(M)} } &  \multirowcell{2}{ {FLOPs} \\ {(G)} }  & \multirow{2}{*}{ {FPS} }  \\ 
\cmidrule(lr){4-5} \cmidrule(lr){6-8}
      &             &          & None           & Full          & S        & W        & G        \\ \midrule
 BSL & ResNet50  & Not Shared &  82.1              &   87.1            & 88.2         & 83.0         & 78.4    &  81 & 236 & {14.2} \\
 {SPTS~\cite{Peng2021SPTSST}} & Swin-B      & Native Shared & 82.5           & 87.3          & 88.5     &  83.2    & 78.7    & 108  & 317  &  {0.6} \\
 {UNITS~\cite{kil2023towards}} & Swin-B      & Native MoE &  83.2   & 88.0       & 88.8    & 83.6     &  79.2   & 128  & 349  & {3.1} \\
OmniParser~\cite{omniparser} &  Swin-B   & Not Shared   & 84.0           & 88.9          & 89.6 & 84.5 & 79.9   &  144 & 436  & {2.6} \\
\midrule
\ourmodel &  Swin-B   & Token Router   & \textbf{84.3}           & \textbf{89.5 }         & \textbf{89.9} & \textbf{84.5} & \textbf{80.6 }  & 110 & 320 & {8.2} \\

\bottomrule
\end{tabular}
\end{adjustbox}

\label{table:abbackbone}
\end{table}

\mypara{Ablating Architectural Designs and Computational Complexity.}
We conducted a comparative analysis of different architectural decoder designs in~\cref{table:abbackbone} to evaluate the effect of decoder weight-sharing and routing mechanisms. The naive shared decoder (Row 2) led to performance degradation on text spotting tasks, indicating conflicts among the subtasks of decoding center points, polygons, and content. Replacing it with our proposed token-router-based decoder mitigates these conflicts through explicit expert selection. \ourmodel achieves an average performance gain of +1.72\% over the naive shared decoder and +0.38\% over OmniParser~\cite{omniparser}, confirming the benefit of structured expert activation. Compared with the native learnable-router MoE decoder (Row 3), our token-router decoder reduces parameter count by 14.1\% and achieves higher inference speed (FPS 8.2 vs. 3.1) while preserving accuracy. Moreover, it eliminates the need for gradient-based routing optimization, enhancing task synergy and simplifying deployment. Our decoder further reduces the total model size by 23.6\% compared to the original OmniParser and improves speed (FPS 8.2 vs. 2.6), supporting its efficiency across VsTP tasks. Additionally, we tested two visual backbone ResNet50 and Swin-B, with Swin-B delivering superior results, reinforcing its suitability for visually-situated text parsing tasks.

\begin{table}[htbp]
\centering
\caption{\textbf{Ablation of decoder length for the table recognition task on PubTabNet datasets.} S-Decoder Len. and C-Decoder Len. refer to the length of the \pointsdecoder for structured points sequence and content sequence, respectively.}
   \begin{adjustbox}{max width=0.47\textwidth}
   \begin{threeparttable}
\begin{tabular}{lcccccc}
\toprule
  Methods                         & S-Decoder Len.      & C-Decoder Len.      & S-TEDS & TEDS & FPS  \\
 \midrule
  Donut~\cite{kim2022donut} & - & - & 25.2 & 22.7 & 0.8 \\
  \midrule
  \multirow{3}{*}{\ourmodel}  & 1,124            & 200             & 89.9  & 88.2 & 2.1 \\
  & 1,500            & 200             & \textbf{90.5} & 88.9  & 1.7 \\
  & 2,000            & 300             & \textbf{90.5}  & \textbf{89.0} & 1.3 \\
\bottomrule
\end{tabular}
   \end{threeparttable}
   \end{adjustbox}

\label{tab:ptn_var_dec_len}
\end{table}

\mypara{Ablating Decoder Length.}
In \cref{tab:ptn_var_dec_len}, we conduct an ablation study on decoder lengths for end-to-end table recognition. Due to GPU constraints, Donut's maximum sequence length is set to 4,000, while our model achieves superior results with a length of only 1,500. Despite using a shorter sequence length, our model demonstrates higher efficiency, achieving an average inference speed of 2.1 FPS, which is nearly three times faster than Donut's 0.8 FPS.
Training an end-to-end model like Donut, which directly uses the complete HTML sequence (including cell text) as ground-truth, is challenging for lengthy tables, often causing error accumulation and attention drift.
In contrast, our modularized architecture separates the table's HTML into two stages: structured points sequence generation and cell text sequence generation, which effectively mitigates these issues and alleviates sequence length limitations, enabling end-to-end table recognition.
Furthermore, increasing the length of the structured points sequence in the \pointsdecoder from 1,500 to 2,000 yields no improvement in S-TEDS, with a slight gain in TEDS when increasing the content sequence length from 200 to 300. 
In practice, choosing the decoder length involves a trade-off between performance and efficiency.

\mypara{Ablating SPOT Prompting on Edge Cases.}
To evaluate the robustness of SPOT prompting under complex real-world scenarios, we conduct an ablation study on  SignaTR6K~\cite{Gholamian2023SignaTR6K} dataset, which includes cluttered layouts with overlapping printed and handwritten text. As shown in~\cref{table:abl_spot_edge_case}, applying a native text-spotting SPOT without layout semantics results in a performance drop (mean F1 of 69.8\%), indicating difficulty in entangled visual structures. Incorporating layout-aware SPOT significantly boosts F1 scores, especially for handwritten text (HT), improving from 61.5\% to 79.3\%. This demonstrates the effectiveness of layout semantics in guiding the model under challenging settings. However, the performance gap between printed and handwritten text reveals remaining limitations and leaves it for future work.

\begin{table}[htbp]
   \centering
      \caption{{\bf Ablation SPOT prompting on SignaTR6K~\cite{Gholamian2023SignaTR6K} under cluttered and overlapping document layouts, including Printed Text (PT) and Handwritten Text (HT) scenarios.} 
      }
   \begin{adjustbox}{max width=0.46\textwidth}
   \begin{threeparttable}
     \centering
     \begin{tabular}{lcccc}
     \toprule
     \multirow{2}{*}{Methods} & \multirow{2}{*}{Training and Test Data} & \multicolumn{3}{c}{F1 (\%)}  \\
      \cmidrule(lr){3-5} 
      &    & PT & HT & Mean  \\
      \midrule 
        MFM~\cite{Gholamian2023SignaTR6K}      & SignaTR6K    & 79.4    & 68.7 & 74.0   \\
        \midrule
        Native Text Spotting SPOT     & SignaTR6K & 78.2  & 61.5  & 69.8 \\
        + Layout Analysis SPOT                & SignaTR6K &  \rankfirst{84.9}     &  \rankfirst{79.3}  &  \rankfirst{82.1}   \\
        
        \bottomrule
     \end{tabular}
     
   \end{threeparttable}
   \end{adjustbox}

   \label{table:abl_spot_edge_case}
\end{table}

\mypara{Qualitative Results.} We present qualitative results for four tasks in~\cref{fig:visualization}, demonstrating the effectiveness of \ourmodel across diverse scenarios. 
For text spotting, our model accurately detects and recognizes curve texts, vertical texts, and artistic texts even in challenging scenarios. Although some detections are slightly imprecise, the recognition results remain fully accurate. 
In KIE, our model effectively localizes and recognizes text while successfully extracting entity information, showcasing its ability to handle structured content extraction tasks.
For table recognition, we present difficult cases, including spanning cells, borderless tables, and multi-line cell content. Our method accurately localizes cell centers using the structured points sequence, enabling reliable table structure reconstruction.
In layout analysis, the results illustrate the model’s ability to group text at the word, line, and paragraph levels, effectively capturing document structures.

\begin{table*}[htbp]
\centering
\caption{\textbf{Quantitative accuracy of text spotting of cooperating structured-points-of-thought prompting with existing multimodal large language model methods.} The `ICDAR 2015', `Total-Text', and `CTW1500' datasets do not use a specific vocabulary for evaluation. Following TextMonkey~\cite{textmonkey}, we use the transcription-based metric `Trans' and point-based metric `Pos' to evaluate methods. * indicates that word-level `Pos' metric is reported for CTW1500. $^\dagger$ indicates the use of additional data from the first stage SFT data of TextMokey~\cite{textmonkey}, with a size of 400k.}
\begin{adjustbox}{max width=0.98\textwidth}
\begin{tabular}{lllllllll}
\toprule
\multirow{2}{*}{}                                                                & \multirow{2}{*}{Method}              & \multirow{2}{*}{Dataset}    & \multicolumn{2}{c}{ICDAR 2015} & \multicolumn{2}{c}{Total-Text} & \multicolumn{2}{c}{CTW1500} \\
\cmidrule(lr){4-5} \cmidrule(lr){6-7} \cmidrule(lr){8-9}
                                                                                 &                                      &                             & Trans       & Pos        & Trans      & Pos       & Tans       & {Pos*}        \\
\midrule                                      
\multirow{5}{*}{\begin{tabular}[c]{@{}l@{}}Specialist\\      Model\end{tabular}} & TOSS~\cite{toss}                                 & -                           & -           & 47.1       & -          & 61.5      & -          & {60.3}       \\
                                                                                 & TTS~\cite{Kittenplon2022TowardsWT}                                  & -                           & -           & 70.1       & -          & 75.1      & -          & -          \\
                                                                                  
                                                                                 & SPTS v2~\cite{liu2023spts}                              & -                           & 55.6        & \textbf{72.6}       & 64.7       & {75.5}      & 55.6       & {75.8}       \\
                                                                                  & WeCromCL+SPTS~\cite{Wu2024WeCromCLWS}                           & -                           & 59.7        & -       & 63.2       & -      & -       & -       \\
                                                                                 & InstructOCR~\cite{Duan2024InstructOCRIB}                           & -                           & 80.6        & -       & 83.4       & -      & -       & -       \\
\midrule
\multirow{3}{*}{MLLM-based}                                                       & Monkey~\cite{monkey}                               & -                           & 43.5        & -          & 48.8       & -         & -          & -          \\
                                                                                  
                                                                                  & StrucTextv3-1.8B~\cite{Lyu2024StrucTexTv3AE}                           & TIM-30M                           & 62.4        & 69.5       & -       & -      & -       & -       \\
                                                                                 & TextMonkey~\cite{textmonkey}                           & 400k                           & 66.9        & 41.6       & 78.2       & 57.8      & 82.1       & {70.3}       \\

\midrule
\multirow{7}{*}{Ours}                                                            & \multirow{7}{*}{InternVL1.5-2B~\cite{internvl15}}                      & {R440k}                         & 62.0        & -          & 71.8       & -         & -          & -          \\
                                                                                 &                      & {R980k}                        & 61.7        & -          & 77.5       & -         & -          & -          \\
 & & R440k+TS180k-N-SPOT                 & 75.6        & 56.1       & 79.9       & 62.8      & 85.2       & 71.7       \\
                                                                                 &                                      & R440k+TS380k-N-SPOT                 & 80.2        & 67.5       & 83.2       & 70.9      & 85.9       & 75.8       \\
                                                                                 &                                      & R440k+TS380k-S-SPOT         & 76.2        & 55.7       & 81.1       & 66.3      & 85.0       & 71.6       \\
                                                                                 &                                      & R440k+TS380k-L-SPOT                 & 79.1        & 65.5       & 82.5       & 68.8      & 85.8       & 75.9       \\
                                                                                 &                                      & R980k+TS380k-N-SPOT$^\dagger$ & 80.7        & 69.1       & 84.2       & 73.3      & 86.2       & 76.3      \\
\midrule

\multirow{6}{*}{Ours}                                                            & \multirow{6}{*}{Mini-Monkey-2B\cite{Huang2024MiniMonkeyMA}}     & {-}                           & 46.2        & -          & 50.5       & -         &  43.1         & -          \\
 & & R440k+TS180k-N-SPOT                 & 76.5        & 57.2       & 80.2       & 63.4      & 86.4       & 72.6       \\
                                                                                 &                                      & R440k+TS380k-N-SPOT                 & 82.4        & 68.4       & 84.4       & 72.5      & 86.8       & 77.3       \\
                                                                                 &                                      & R440k+TS380k-S-SPOT         & 77.5        & 57.5       & 82.6       & 67.6      & 86.1       & 71.9       \\
                                                                                 &                                      & R440k+TS380k-L-SPOT                 & 80.4        & 67.2       & 83.8       & 69.4      & 86.6       & 77.8       \\
                                                                                 &                                      & R980k+TS380k-N-SPOT$^\dagger$ & 81.6        & 69.8       & 84.9       & 75.4      & 87.9       & 78.4      \\

\midrule

\multirow{6}{*}{Ours}  & \multirow{6}{*}{Qwen2-VL-2B~\cite{Qwen2-VL}}                           & {-}                           & 48.3        & -          &  55.7       & -         &  46.4         & -          \\

 & & R440k + TS180k-N-SPOT                 &    77.1     &   59.2  &  81.3     &   65.6    &    87.7    &   73.7   \\
                       &                                      & R440k+TS380k-N-SPOT              &    {83.4}     &    69.8 &   {85.2}    &    73.4   &    {88.0}    &    {78.5}  \\    
                        &                                      & R440k+TS380k-S-SPOT       &    78.6    &    59.9 &    81.6   &     69.2  &     87.9   &   72.4   \\   
                      &                                      & R440k+TS380k-L-SPOT               &     82.5    &    68.6 &    83.5   &    69.7   &    87.0    &   77.8   \\   
                      &                                      & R980k+TS380k-N-SPOT$^\dagger$   &  \ranksecond{83.8}  & {70.0}  &    \ranksecond{85.8}   &    \textbf{75.9}   &     \ranksecond{88.4}   &    \ranksecond{78.7}  \\                                                                 
\midrule

\multirow{2}{*}{{Ours}}  & \multirow{2}{*}{{InternVL2.5-2B~\cite{Chen2024InternVl25}}} & -      &    63.6     &   45.3  &  70.8     &   51.5    &    74.2    &  50.6    \\
                       &                                      & R440k+TS380k-N-SPOT              &    {81.5}     &    68.6 &   {84.7}    &    72.6   &    {87.1}    &    {77.4}  \\ 

\midrule

\multirow{2}{*}{{Ours}}  & \multirow{2}{*}{{Qwen2.5-VL-3B~\cite{Bai2025Qwen25VLTR}} } & -      &    68.5     &   50.4  &  74.7     &   57.1    &    76.6    &   57.9   \\
                       &                                      & R440k+TS380k-N-SPOT              &    \rankfirst{84.8}     &    \ranksecond{72.3} &   \rankfirst{86.3}    &    \ranksecond{75.8}   &    \rankfirst{89.8}    &    \rankfirst{79.4}  \\                                                                      

\bottomrule
\end{tabular}
\end{adjustbox}

\label{table:mllm_on_spotting}
\end{table*}

\subsection{SPOT Applied to Existing MLLMs}
\mypara{{Text Spotting.}} To further validate the generalizability of our structured-points-of-thought prompting, we integrate SPOT with existing multimodal large language models and evaluate their performance on text spotting. Following TextMonkey~\cite{textmonkey}, we adopt the transcription-based metric `Trans' and the point-based metric `Pos'. As shown in~\cref{table:mllm_on_spotting}, we experiment with five representative MLLMs: InternVL1.5-2B~\cite{internvl15}, InternVL2.5-2B~\cite{Chen2024InternVl25}, Mini-Monkey-2B~\cite{Huang2024MiniMonkeyMA}, Qwen2-VL-2B~\cite{Qwen2-VL}, and Qwen2.5-VL-3B~\cite{Bai2025Qwen25VLTR}. 
Without SPOT prompting, using only the R440k or R980k datasets with InternVL1.5-2B fails to outperform specialist systems such as InstructOCR~\cite{Duan2024InstructOCRIB} or document-oriented MLLM baselines like TextMonkey~\cite{textmonkey} on ICDAR2015 and Total-Text. Although StrucTextv3 achieves strong performance through a 30M-scale text spotting corpus (TIM-30M), our setting \texttt{R980k+TS380k-N-SPOT} still attains competitive Pos scores on ICDAR2015, demonstrating the effectiveness of task-aware prompting even with far less training data.
With the integration of SPOT prompting, all five MLLMs exhibit consistent and substantial gains across ICDAR2015, Total-Text, and CTW1500. The improvements hold for both Trans and Pos metrics, and SPOT-enhanced models surpass specialist methods (e.g., InstructOCR) and prior MLLM-based approaches (e.g., TextMonkey, StrucTextv3). We further include InternVL2.5-2B and Qwen2.5-VL-3B as stronger contemporary baselines in~\cref{table:mllm_on_spotting}, and SPOT continues to yield clear improvements over these state-of-the-art MLLMs, reinforcing its robustness and scalability.
This superior performance is attributed to its ability that SPOT prompting effectively anchors the model’s inference to spatially grounded text center points. Additional gains are obtained when incorporating larger-scale spotting SFT data (400k samples) from TextMonkey~\cite{textmonkey}. Overall, SPOT prompting significantly enhances text localization and recognition capabilities across a diverse set of modern MLLMs, validating its general applicability.

\begin{table}[htbp]
   \centering
      \caption{{\bf Hallucination of the resulting model Qwen2.5VL with SPOT prompting on HallusionBench~\cite{Guan2023HallusionbenchAA}. }
      }
   \begin{adjustbox}{max width=0.46\textwidth}
   \begin{threeparttable}
     \centering
    
     \begin{tabular}{lc}
     \toprule
     \multirow{1}{*}{Methods} & \multirow{1}{*}{HallusionBench (\%)}   \\
      
      \midrule 
       
        Qwen2.5VL-3B~\cite{Bai2025Qwen25VLTR}     & 63.8 \\
        + Text Spotting SPOT Prompting     & 68.5  \\
        
        \bottomrule
     \end{tabular}

   \end{threeparttable}
   \end{adjustbox}

   \label{table:hallucination_of_res_model}
\end{table}

\mypara{Hallucination of the Resulting Model.}
To validate the effectiveness of SPOT prompting in mitigating hallucination within MLLMs, we evaluate Qwen2.5-VL-3B on HallusionBench~\cite{Guan2023HallusionbenchAA}, a benchmark specifically designed for factuality assessment in vision-language tasks. As shown in~\cref{table:hallucination_of_res_model}, incorporating SPOT prompting improves hallucination resistance, increasing accuracy from 63.8\% to 68.5\%. This result provides quantitative support for our claim that SPOT enhances grounding and reduces hallucinated outputs by anchoring model predictions to structured visual semantics.

\mypara{The Necessity of Equipping SPOT with Existing MLLMs.} While \ourmodel has already achieved high performance in visually-situated text parsing (as shown in \cref{table:textspotting}) and SPOT prompting has demonstrated notable improvements for MLLMs in text localization (as seen in \cref{table:mllm_on_spotting}), a performance gap remains between MLLMs and \ourmodel. Nevertheless, enhancing the native visual text parsing capabilities of MLLMs is crucial. Unlike directly using existing tools such as \ourmodel, enabling MLLMs to independently perform these tasks fosters a more integrated, end-to-end multimodal reasoning system, reducing external dependencies and improving overall efficiency.
Without native text perception, the model would need to rely on external OCR tools, leading to potential delays, information loss, or hallucination. A text-aware MLLM, however, can directly detect and read the relevant text within the document and provide accurate answers efficiently, making it more suitable for real-world applications.
Thus, enhancing MLLMs' text perception through SPOT prompting is only an initial step. Future work will explore more effective approaches to natively improve MLLMs' visual text parsing capabilities, enabling them to perform robustly in complex, real-world multimodal tasks.

\mypara{Ablating SPOT Prompting Length.} We conducted ablation studies with three different lengths of SPOT prompting, referred to as normal SPOT (N-SPOT), short SPOT (S-SPOT), and long SPOT (L-SPOT), to fine-tune the MLLM. As shown in \cref{table:mllm_on_spotting}, we found that long SPOT prompting, which includes additional steps including detection and recognition prompting, tends to cause a slight performance degradation, likely due to increased complexity and longer inference sequences~\cite{Jin2024TheIO}. On the other hand, short SPOT prompting, which omits the intermediate generation of structured points sequence, results in a performance drop, highlighting the importance of intermediate reasoning steps in the SPOT framework.

\mypara{Key Information Extraction.}
As shown in~\cref{table:kie_spot}, SPOT prompting consistently enhances MLLM performance on entity-level key information extraction across both CORD and SROIE datasets. For example, InternVL2.5-2B improves from 74.5\% to 83.7\% F1 on CORD and from 71.9\% to 77.8\% F1 on SROIE after integrating SPOT. Similarly, Qwen2.5-VL-3B shows substantial gains, reaching 84.8\% and 83.8\% F1 on CORD and SROIE, respectively. These results confirm SPOT’s ability to inject structural guidance into MLLMs, yielding better field-level recognition and layout-aware understanding for document-level entity parsing.

\begin{table}[htbp]
   \centering
      \caption{{\bf Key information extraction results of cooperating SPOT prompting with existing MLLM methods. }
      }
   \begin{adjustbox}{max width=0.46\textwidth}
   \begin{threeparttable}
     \centering
     
     \begin{tabular}{lcccc}
     \toprule
     \multirow{2}{*}{Methods} & \multicolumn{2}{c}{CORD} & \multicolumn{2}{c}{SROIE}  \\
      \cmidrule(lr){2-3} \cmidrule(lr){4-5}
      &   F1 & Acc & F1 & Acc  \\
      \midrule 
        InternVL2.5-2B~\cite{Chen2024InternVl25}      & 74.5    & 78.2     & 71.9 & 76.3   \\
        InternVL2.5-2B+N-SPOT                     & 83.7    &   81.9     & 77.8 & 85.7   \\
        \midrule
        Qwen2.5-VL-3B~\cite{Bai2025Qwen25VLTR}     & 77.2 & 82.6  & 76.1  & 82.8 \\
        Qwen2.5-VL-3B+N-SPOT                & \rankfirst{84.8} &  \rankfirst{86.4}     &  \rankfirst{83.8}  &  \rankfirst{89.1}   \\
        
        \bottomrule
     \end{tabular}

   \end{threeparttable}
   \end{adjustbox}

   \label{table:kie_spot}
\end{table}

\mypara{Table Recognition.} 
As shown in~\cref{tab:ptn_and_ftn_spot}, SPOT prompting enhances the table recognition capabilities of MLLMs on both PubTabNet and FinTabNet datasets. For InternVL2.5, SPOT improves S-TEDS by +8.9\% and TEDS by +7.8\% on PubTabNet, while also boosting FinTabNet scores. Qwen2.5-VL, which performs strongly even without SPOT, still benefits significantly, achieving superior results with 82.5\% S-TEDS and 80.1\% TEDS on FinTabNet. These results demonstrate SPOT's consistent effectiveness in guiding MLLMs to better capture structural table semantics.

\begin{table}[htbp]
\centering
\caption{\textbf{Table recognition results of cooperating SPOT prompting with existing MLLM methods. } 
}
   \begin{adjustbox}{max width=0.47\textwidth}
   \begin{threeparttable}

\begin{tabular}{lccc}
\toprule
 \multicolumn{3}{c}{PubTabNet (PTN)}       \\
 \midrule
  Methods                         & S-TEDS & TEDS  \\
 \midrule
InternVL2.5-2B~\cite{Chen2024InternVl25} &  68.4 & 62.7 \\ 
InternVL2.5-2B+N-SPOT      &  77.3  & 70.5 \\ 
\midrule
Qwen2.5-VL-3B~\cite{Bai2025Qwen25VLTR} &  76.9 &  73.8 \\ 
Qwen2.5-VL-3B+N-SPOT        &   \textbf{79.8} &  \textbf{76.2} \\ 
\toprule

 \multicolumn{3}{c}{FinTabNet (FTN)}       \\
 
 \midrule
  Methods                             & S-TEDS & TEDS  \\
 \midrule
InternVL2.5-2B~\cite{Chen2024InternVl25}    &  67.4  &  63.2 \\ 
InternVL2.5-2B+N-SPOT      &  70.2   & 69.6 \\ 
\midrule
Qwen2.5-VL-3B~\cite{Bai2025Qwen25VLTR}    &  78.1   &  75.7 \\ 
Qwen2.5-VL-3B+N-SPOT           & \textbf{ 82.5 }  & \textbf{80.1} \\

\bottomrule
\end{tabular}

   \end{threeparttable}
   \end{adjustbox}

\label{tab:ptn_and_ftn_spot}
\end{table}

\begin{table}[htbp]
   \centering
      \caption{{\bf Layout analysis results of cooperating SPOT prompting with existing MLLM methods. }
      }
  \setlength\tabcolsep{1.3pt}
   \begin{adjustbox}{max width=0.49\textwidth}
   \begin{threeparttable}
     \centering
     
     \begin{tabular}{lccc}

     \toprule
     \multicolumn{4}{c}{HierText Validation Set}       \\
     \midrule
     \multirow{2}{*}{Methods} & \multicolumn{1}{c}{Word-level} & \multicolumn{1}{c}{Line-level} & \multicolumn{1}{c}{Paragraph-level}  \\
      \cmidrule(lr){2-2} \cmidrule(lr){3-3} \cmidrule(lr){4-4}
      &  PQ & PQ & PQ  \\
      \midrule 
InternVL2.5-2B~\cite{Chen2024InternVl25} &  49.5 & 55.2  & 46.8 \\ 
InternVL2.5-2B+N-SPOT      & 53.1    & 59.6  & 53.1 \\ 
\midrule
Qwen2.5-VL-3B~\cite{Bai2025Qwen25VLTR} & 52.7 &  57.3  & 49.6 \\ 
Qwen2.5-VL-3B+N-SPOT        &   \textbf{58.4} &  \textbf{60.7} &  \textbf{53.8} \\

     \toprule
     \multicolumn{4}{c}{HierText Test Set}       \\
     \midrule
     \multirow{2}{*}{Methods}  & \multicolumn{1}{c}{Word-level} & \multicolumn{1}{c}{Line-level} & \multicolumn{1}{c}{Paragraph-level}  \\
      \cmidrule(lr){2-2} \cmidrule(lr){3-3} \cmidrule(lr){4-4}
      &   PQ & PQ & PQ  \\
      \midrule 
InternVL2.5-2B~\cite{Chen2024InternVl25} &  51.3 &  55.8 & 42.9 \\ 
InternVL2.5-2B+N-SPOT      &  58.0  & 59.3  & 49.6 \\ 
\midrule
Qwen2.5-VL-3B~\cite{Bai2025Qwen25VLTR} & 57.2 &  58.8 & 50.7 \\ 
Qwen2.5-VL-3B+N-SPOT        &  \textbf{60.7}  & \textbf{63.5} & \textbf{53.8} \\

        \bottomrule
     \end{tabular}
 
   \end{threeparttable}
   \end{adjustbox}

   \label{table:layout_analysis_spot}
\end{table}

\mypara{Layout Analysis.} 
To validate SPOT prompting for structural understanding, we evaluate its integration with MLLMs on the HierText dataset. As shown in~\cref{table:layout_analysis_spot}, SPOT consistently improves layout parsing across word-level, line-level, and paragraph-level metrics. On the test set, InternVL2.5 with SPOT improves from 51.3\% to 58.0\% (word PQ), while Qwen2.5-VL achieves further gains from 57.2\% to 60.7\%, demonstrating the effectiveness of SPOT in enhancing layout-aware parsing.

\subsection{Limitation}
Although \ourmodel achieves superior performance in unified visually-situated text parsing, its understanding and reasoning capabilities remain limited. Integrating reinforcement learning techniques, such as GRPO~\cite{deepseekai2025deepseekr1simple}, is one of the promising directions for our future work to enhance the model’s reasoning ability, interpretability, and comprehension in document-oriented multimodal tasks.

\section{Conclusion}\label{sec:conclusion}
In this paper, we proposed a general-purpose visual text parsing framework, \ourmodel, which unifies the tasks of text spotting, key information extraction, table recognition, and layout analysis within a visually-situated text parsing context. This is achieved through a two-stage decoding procedure via SPOT prompting, where structured points act as an adapter to bridge different tasks.
To further enhance pre-training effectiveness across all tasks, we introduced two specialized pre-training strategies, enabling the \pointsdecoder to learn complex structures and relationships among visually-situated texts, thereby improving overall performance.
The proposed \ourmodel achieves state-of-the-art or highly competitive performance on standard benchmarks for all four tasks, outperforming or matching specialist models that rely on task-specific designs. Additionally, the generality of SPOT prompting is demonstrated through its superior text parsing capabilities when applied to existing MLLMs, highlighting its effectiveness in improving the text understanding abilities of large multimodal models.
We hope this work serves as a foundation for future advancements toward building generalized unified frameworks for document understanding.

\ifCLASSOPTIONcaptionsoff
  \newpage
\fi

{
\bibliographystyle{IEEEtran}
 \bibliography{strings_abbr, references}

}

\vfill 
\end{document}

\title{OmniParser V2: Structured-Points-of-Thought for Unified Visual Text Parsing and Its Generality to Multimodal Large Language Models\\--- Supplemental Material ---}

\maketitle

\section{Datasets and Evaluation Metrics}

\subsection{Text Spotting}

\mypara{Datasets.}
We conduct experiments on three popular scene text datasets, Total-Text, ICDAR 2015, and CTW1500~\cite{liu2019curved}. 
Total-Text is mainly for arbitrary-shaped text detection and spotting evaluation, consisting of 1255 training images and 300 testing images with word-level polygon annotations. 
The ICDAR 2015 dataset contains 1000 training images and 500 testing images, annotated with quadrilateral bounding boxes. 
CTW1500 is another benchmark for curved text detection and recognition, which is annotated at text-line level, including 1000 training images and 500 testing images.

\mypara{Evaluation Metrics.}
For Total-Text and CTW1500, we report the end-to-end recognition results over two lexicons: ``None'' and ``Full''. ``None'' means that no lexicons are provided, and ``Full'' lexicon provides all words in the test set.
For ICDAR 2015, we report results over three lexicons: ``Strong'', ``Weak'' and ``Generic''. Strong lexicon provides 100 words that may appear in each image. Weak lexicon provides words in the whole test set, and generic lexicon provides a 90k vocabulary.

\subsection{Key Information Extraction}

\mypara{Datasets.}
We evaluate our model's performance on two commonly used benchmark datasets for KIE task: CORD~\cite{park2019cord} and SROIE~\cite{huang2019icdar2019}. 
CORD~\cite{park2019cord} consists of 30 labels across 4 categories.
It has 1,000 receipt samples. The train, validation, and test splits contain 800, 100, and 100 samples respectively. 
The SROIE dataset~\cite{huang2019icdar2019} comprises a training set with 626 receipts and a test set with 347 receipts. Each receipt in the dataset contains four predefined entities, namely: ``company'', ``date'', ``address'', and ``total''. Annotations in the dataset provide segment-level bounding boxes for the text regions and their corresponding transcriptions. 

\mypara{Evaluation Metrics.}
Following~\cite{kim2022donut}, two evaluation metrics are used to evaluate the performance: field-level F1 measure and tree-edit-distance-based accuracy.
The field-level F1 score checks whether each extracted field corresponds exactly to its value in the ground truth.

\subsection{Table Recognition}
\mypara{Datasets.} 
Given our model's dual prediction of table logical structures (with cell bounding box central points) and cell content, datasets lacking annotations for both cell content and corresponding bounding boxes, as well as those using metrics incompatible with our approach, are excluded from evaluation. For model assessment, PubTabNet (PTN)~\cite{EDD} and FinTabNet (FTN)~\cite{GTE} are selected. PubTabNet has 500,777 training images and 9,115 validation images, featuring diverse structures from scientific documents. Our model is evaluated on the validation set due to the lack of public annotations for the test set. FinTabNet comprises 112k single-page PDFs with 92,000 cropped training images and 10,656 testing images. 

\mypara{Evaluation Metrics.} For evaluation, we utilized Tree-Edit-Distance-based Similarity (TEDS)~\cite{EDD}. TEDS comprehensively evaluates table similarity, considering both structural and cell content aspects in HTML format. The metric represents the HTML table as a tree, and the TEDS score is computed through the tree-edit distance between the ground truth and predicted trees. In addition to overall results, we also provide S-TEDS results, focusing exclusively on the structural aspects and ignoring cell content.

\subsection{Layout Analysis}
\mypara{Datasets.} We evaluate layout analysis on HierText~\cite{long2022towards}, which consists of 8,281 training images, 1,724 validation images, and 1,634 testing images. It has dense and small text instances from various sources, including scene, designed, printed, and handwritten texts. It provides hierarchical word, text-line, and paragraph location annotations. Word locations are annotated with polygons while text-line locations are annotated with quadrilateral boxes. Paragraph locations are represented by coarse polygons.

\mypara{Evaluation Metrics.} Following HierText~\cite{long2022towards}, we evaluate word, line, and paragraph grouping tasks using Panoptic Quality (PQ)~\cite{Kirillov2018PanopticS}. The primary evaluation metric is the paragraph-level PQ metric.

\subsection{SPOT-style SFT Dataset for MLLMs}
We used publicly available datasets and data collected from Platypus~\cite{Wang2024PlatypusAG} to construct the SPOT-style supervised fine-tuning (SFT) datasets for multimodal large language models. The settings, sizes, and sources in our curated SPOT-style SFT data models are shown in~\cref{tab:spot_like_sft_data_detail}. Rico~\cite{rico} uses pseudo-labels generated by our internal business's small model for detection and recognition, while all other annotated open-source datasets use raw OCR annotations.

\begin{table}[htbp]
\centering
\caption{\textbf{The settings and number of our curated SFT data.} Num. short for number. }
   \begin{adjustbox}{max width=0.49\textwidth}
   \begin{threeparttable}
\begin{tabular}{lccccl}
\toprule
  Task Type                         & Prompt Setting     & Data Num.  & Data Name  & Data Sources \\
 \midrule
  Text Spotting  &  \makecell[l]{N-SPOT,\\ S-SPOT, \\L-SPOT}       & 181,593    & TS180k  & \makecell[l]{ICDAR2013~\cite{Karatzas2013ICDAR2R},\\ HierText~\cite{long2022towards},\\ TextOCR~\cite{singh2021textocr},\\ SynthText150k~\cite{Liu2021ABCNetVA}} \\
  \midrule
 Text Spotting  &      \makecell[l]{N-SPOT,\\ S-SPOT, \\L-SPOT}         & 389,433            & TS380k & \makecell[l]{ICDAR2013~\cite{Karatzas2013ICDAR2R},\\ HierText~\cite{long2022towards},\\ TextOCR~\cite{singh2021textocr},\\ SynthText150k~\cite{Liu2021ABCNetVA},\\ OpenImageV5 Text~\cite{krylov2021open}}  \\
   \midrule
  Read All Text  &  \makecell[l] {Read all the text \\in the image.}           & 446,702        & R440k & \makecell[l]{DocLayNet~\cite{DocLayNet},\\ HierText~\cite{long2022towards},\\ TextOCR~\cite{singh2021textocr},\\ SynthText150k~\cite{Liu2021ABCNetVA},\\ OpenImageV5 Text~\cite{krylov2021open}, \\MLT2017~\cite{nayef2017icdar2017},\\ COCO-Text~\cite{veit2016coco},\\ D4LA~\cite{Da_2023_ICCV}}  \\
  \midrule
 Read All Text &        \makecell[l] {Read all the text \\in the image.}      & 981,284             & R980k & \makecell[l]{DocLayNet~\cite{DocLayNet},\\ HierText~\cite{long2022towards},\\ TextOCR~\cite{singh2021textocr},\\ SynthText150k~\cite{Liu2021ABCNetVA},\\ OpenImageV5 Text~\cite{krylov2021open}, \\MLT2017~\cite{nayef2017icdar2017},\\ COCO-Text~\cite{veit2016coco},\\  SynthDoG\_HW~\cite{kim2022donut},\\ PubLayNet~\cite{zhong2019publaynet},\\ LAION-OCR~\cite{Schuhmann2022LAION5BAO},\\ D4LA~\cite{Da_2023_ICCV},\\ CTIG-DM~\cite{Zhu2023ConditionalTI},\\ Rico~\cite{rico} } \\

\bottomrule
\end{tabular}
   \end{threeparttable}
   \end{adjustbox}

\label{tab:spot_like_sft_data_detail}
\end{table}

\section{Implementation Details}
\subsection{Spatial-Window Prompting}
Spatial-window prompting comprises two components: fixed mode and random mode. In the fixed mode, the image is divided into grid blocks evenly, such as 3x3 or 2x2. Conversely, in the random mode, the starting point of the spatial window is randomly determined. In order to encompass more texts within the random box, the area of the random box is established to be no less than 1/9 of the original image. To elaborate further, a 30\% probability is assigned for selecting the fixed mode, another 30\% probability for selecting the random mode, and a 40\% probability for the defaulting window to cover the entire image. Following ~\cite{kil2023towards}, we set the bin size of the coordinate vocabulary as 1000. The pseudo-code of spatial-window prompting is shown in the following.

\begin{lstlisting}[language=Python]
import random

# prob for different mode
prob = random.uniform(0, 1)

# quantizing coordinates with n_bins
n_bins = 1000

if prob < 0.4:
    # default window
    start_x, start_y, end_x, end_y = [0, 0, n_bins - 1, n_bins - 1]
elif prob < 0.7:
    # x-axis and y-axis are partitioned into varying numbers of blocks.
    num_xs = [3, 3, 1, 3, 2, 2, 2, 1]
    num_ys = [3, 1, 3, 2, 3, 2, 1, 2]

    total_windows = []
    for num_x, num_y in zip(num_xs, num_ys):
        inter_x = min(int(n_bins / num_x), n_bins - 1)
        inter_y = min(int(n_bins / num_y), n_bins - 1)
        
        for i in range(num_x):
            for j in range(num_y):
                start_x = i*inter_x
                start_y = j*inter_y
                end_x = min(start_x + inter_x, n_bins - 1)
                end_y = min(start_y + inter_y, n_bins - 1)
                total_windows.append([start_x, start_y, end_x, end_y])
    
    start_x, start_y, end_x, end_y = random.choice(total_windows)
else:
    inter = int(n_bins / 3)
    start_x = random.randint(0, inter * 2)
    start_y = random.randint(0, inter * 2)
    rect_w, rect_h = random.randint(inter, n_bins - 1), random.randint(inter, n_bins - 1)
    end_x, end_y = min(start_x + rect_w, n_bins - 1), min(start_y + rect_h, n_bins - 1)

spatial_window_prompt = [start_x, start_y, end_x, end_y]

\end{lstlisting}

\subsection{Table Recognition}

Given a table image, we resize it to 1,024$\times$1,024 pixels. The \pointsdecoder, utilizing the feature vector from the Image Encoder, simultaneously generates pure HTML tags with structural cell point sequences in the same sequence representing the table's logical and physical structures. These structural cell point sequences serve as start-prompting input for the \contentdecoder, which extracts table cell contents in parallel. The final output combines pure HTML tags with cell contents, forming complete HTML sequences faithfully representing the table's structure and content.

\mypara{Datasets.} 
Since our model predicts both the logical structure of tables with cell bounding box central points and cell content, datasets lacking cell content and corresponding bounding box annotations, such as TABLE2LATEX-450K~\cite{deng2019challenges}, TableBank~\cite{li2020tablebank}, UNLV~\cite{shahab2010open}, IC19B2H~\cite{gao2019icdar}, WTW~\cite{long2021parsing} and TUCD~\cite{raja2021visual}, are not suitable for our approach. Similarly, datasets like ICDAR2013Table~\cite{gobel2013icdar}, SciTSR~\cite{chi2019complicated}, and PubTables-1M~\cite{smock2022pubtables}, which provide cell content and content box annotations, employ metrics based on box representations that are incompatible with our point-based format. Consequently, PubTabNet (PTN)~\cite{EDD} and FinTabNet (FTN)~\cite{GTE} are selected for our model evaluation.

\begin{figure}[htbp]
  \centering \includegraphics[width=0.96\linewidth]{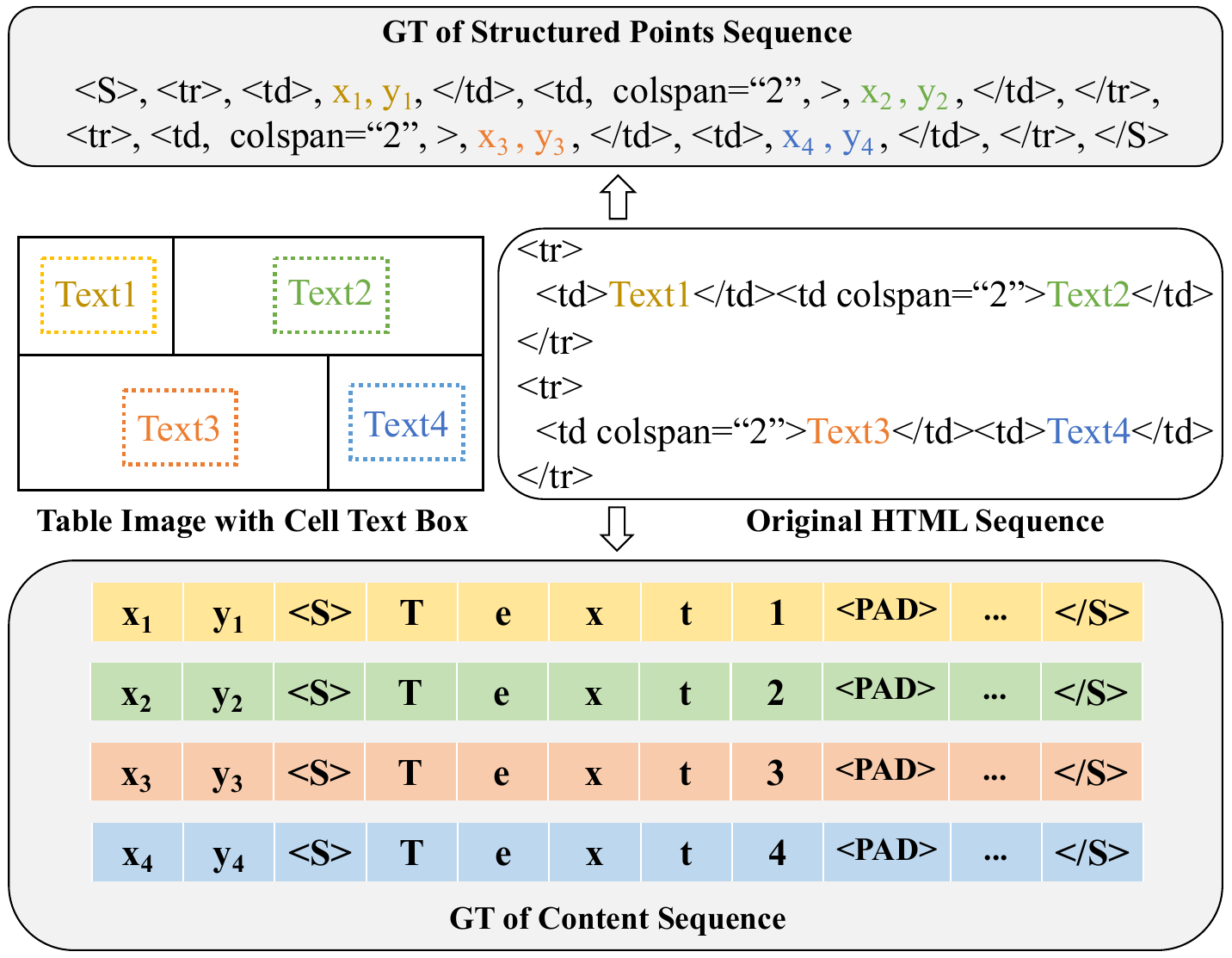}
   \captionsetup{width=0.96\linewidth}
   \caption{\textbf{An example of building training GTs for table recognition task.} We use the center points of each cell text box to build GTs for the \pointsdecoder and the \contentdecoder. If the cell is empty text, the corresponding points in the GTs are left empty as well. }
   \label{fig:gt_table_stru}
\end{figure}

\mypara{GT Generation.} The ground truth pure HTML tags of tables are tokenized into structural tokens. Following the previous works~\cite{TableMaster, VAST}, we use the merged labels to represent a non-spanning cell to reduce the length of the HTML tags. Specifically, we use 
\emph{\textless td\textgreater\textless/td\textgreater} and \emph{\textless td\textgreater[]\textless/td\textgreater} to denote empty cells and non-empty cells, respectively. For a cell spanning multiple rows or columns, the original HTML tags are broken into four tokens: \emph{\textless td}, \emph{colspan=``n"} or \emph{rowspan=``n"}, \emph{\textgreater}, and \emph{\textless/td\textgreater}. We use the first token \emph{\textless td} to represent a spanning cell. In addition, four special symbol categories need to be added: \emph{\textless S\textgreater}, \emph{\textless /S\textgreater}, \emph{\textless PAD\textgreater}, and \emph{\textless UNK\textgreater}, which represent the beginning of a sequence, the end of a sequence, padding symbols, and unknown characters, respectively. For building the GT of \pointsdecoder, we insert center points of each cell text box to corresponding HTML tags. For building the GT of \contentdecoder, we combine each cell text with corresponding center points as a whole sequence where center points can be viewed as a start-prompting input for recognizing text, and each cell text is tokenized at the character level. An example of building a training sequence GT for the \pointsdecoder and the \contentdecoder in the table recognition task is illustrated in~\cref{fig:gt_table_stru}.

\subsection{Layout Analysis}
\begin{figure}[htbp]
  \centering \includegraphics[width=0.98\linewidth]{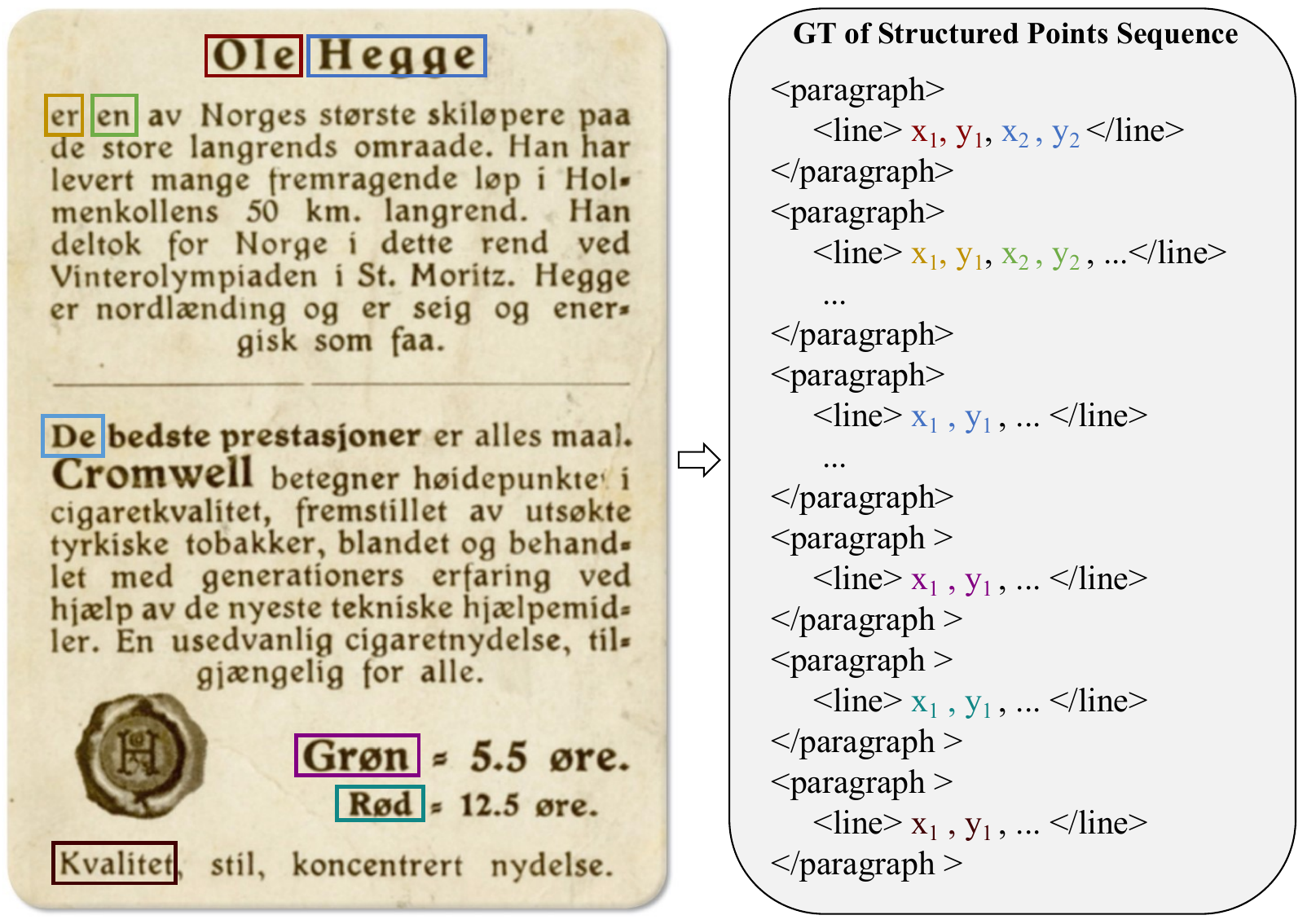}
   \captionsetup{width=0.98\linewidth}
   \caption{\textbf{An Example of building training GTs for layout analysis task.} }
   \label{fig:gt_layout_stru}
\end{figure}

Thanks to the flexible expression of structured sequence in \ourmodel, it is convenient for us to extend it to other OCR-related tasks, such as layout analysis, which aims to group the text in the image into three levels, namely word, line, and paragraph, based on spatial position and semantic relationship. Previous methods~\cite{long2022towards} mainly achieved hierarchical results by clustering based on similarity. In our approach, we distinguish the text belonging to different hierarchical intervals by simply inserting \emph{\textless line\textgreater} and \emph{\textless paragraph\textgreater} structural tags into the sequence of text center points, as shown in~\cref{fig:gt_layout_stru}.

\begin{figure*}[htbp]
  \centering \includegraphics[width=0.95\linewidth]{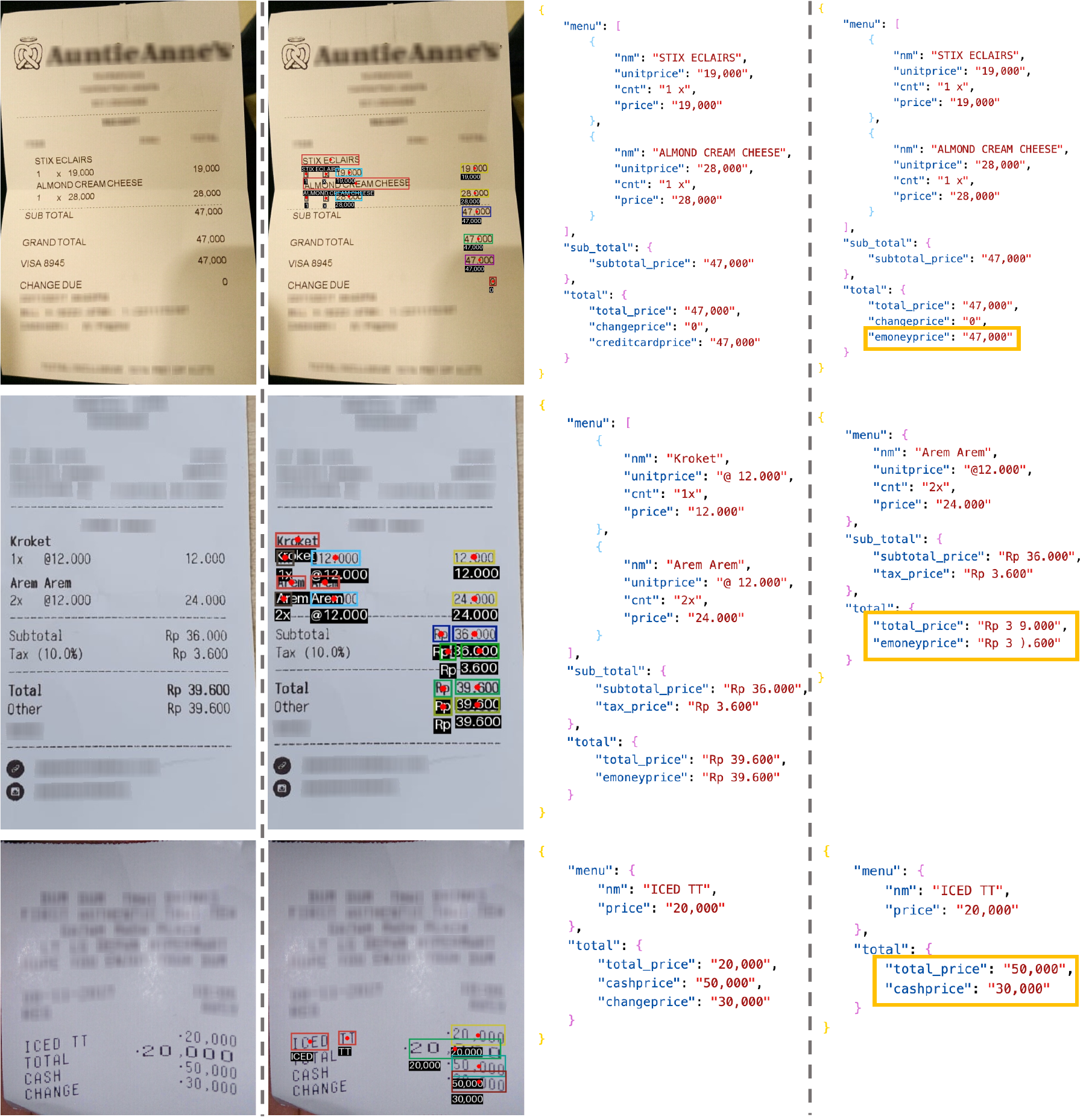}
   \captionsetup{width=0.95\linewidth}
   \caption{\textbf{A comparative analysis of partial results obtained from \ourmodel and Donut on CORD.} The first column depicts the original image, while columns 2 and 3 illustrate our detection results and the corresponding formatted output, respectively. Column 4 showcases the Donut's formatted output. Notably, our model demonstrates superior performance in entity extraction. }
   \label{fig:vs_donut_on_cord}
\end{figure*}

\begin{figure*}[htbp]
  \centering \includegraphics[width=0.95\linewidth]{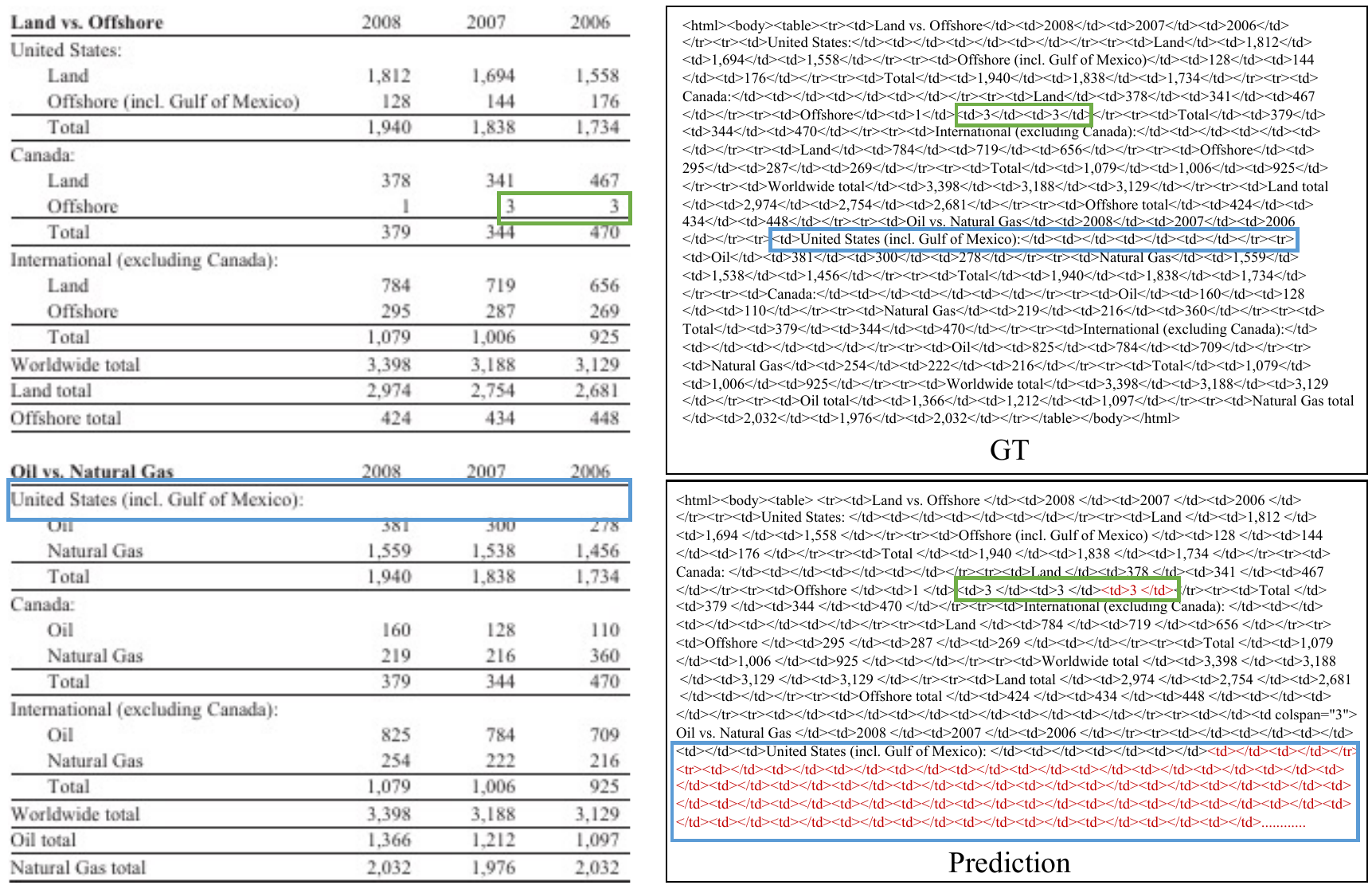}
   \captionsetup{width=0.95\linewidth}
   \caption{\textbf{Illustrative failure case of Donut in table recognition task.} Red text means error predictions. For readability, we only highlight two errors in this example. Due to the lack of point location information, Donut has an attention drift problem, resulting in the prediction of repeated tokens and leading to a high probability of error accumulation in long-sequence scenarios. (The figure is best viewed in color.)}
   \label{fig:donut_table_failure_case}
\end{figure*}

\section{Comparisons with Donut on KIE Task}
As shown in ~\cref{fig:vs_donut_on_cord}, \ourmodel can achieve entity extraction while predicting the location of each entity word.
However, Donut only predicts the structured sequence for entity extraction without any localization ability.
Thus, the absence of direct region supervision during both training and prediction stages often leads to inferior results for entities of the same values (Row 1), repeated entities (Row 2), or entities with explicit trigger names (Row 3).

\section{Training Donut on Table Recognition Task}
We fine-tuned the OCR-free end-to-end model Donut~\cite{kim2022donut} for table recognition on FinTabNet dataset. The ground truth sequence utilized combined HTML tags with table cell text, and we use different training hyper-parameters for adequate verification, as shown in \cref{tab:donut_abla}. Due to GPU memory limitations, we constrained the decoder's max length in Donut to 4,000. Note that the original HTML sequence max lengths for PubTabNet and FinTabNet are 8,722 and 8,035, respectively. 
For long sequence prediction tasks such as table recognition, training an end-to-end model like Donut with combined HTML stages, including cell text, is non-trivial.
There is a high probability of error accumulation and attention drift in long-sequence scenarios leading to the inferior performance of Donut for table recognition. An illustrative example of a failure case for Donut in table recognition task is shown in~\cref{fig:donut_table_failure_case}. 
Specifically, due to the lack of region supervision, the end-to-end model Donut has demonstrated an attention drift problem, resulting in the prediction of repeated tokens and leading to a high probability of error accumulation in long-sequence scenarios. In contrast, \ourmodel decomposes the location-aware structured points sequence and cell text recognition generation, alleviating the issues of attention drift and error accumulation.

\begin{table}[htbp]
\centering
\caption{\textbf{Comparisons of different training hyper-parameters of Donut on FinTabNet datasets.} LR is short for learning rate.}
   \begin{adjustbox}{max width=0.47\textwidth}
   \begin{threeparttable}
\begin{tabular}{lccccc}
\toprule
 
  Methods                            & LR      & Epoch      & S-TEDS & TEDS  \\
 \midrule

\multirow{5}{*}{Donut~\cite{kim2022donut} }                                                   & 3e-5            & 20               &  22.2  & 17.2  \\
                                                 & 3e-5            & 40               &  26.2  & 20.0  \\
                                                   & 1e-4            & 40               &  30.7  & 29.1  \\
                                                    & 1e-3            & 40               &  41.7  & 40.5  \\
                                                   & 1e-3            & 100               &   41.9 & 41.2  \\
 \midrule
\multirow{1}{*}{\ourmodel (ours)}  &   -          & -             & \textbf{93.2}  & \textbf{90.5} \\

\bottomrule
\end{tabular}
   \end{threeparttable}
   \end{adjustbox}
   
\label{tab:donut_abla}
\end{table}

\section{More Results and Analysis}

\subsection{Evaluation on Additional Layout Analysis Datasets}
As shown in \cref{table:more_layout_ana_res}, we have extended our experiments to include PubLayNet~\cite{zhong2019publaynet}, DocLayNet~\cite{DocLayNet}, and M6Doc~\cite{Cheng2023M6DocAL}, three widely adopted benchmarks in layout analysis. The results show that \ourmodel achieves strong performance across all three datasets, with F1 scores of 97.4\% on PubLayNet, mAP of 77.3\% on DocLayNet, and mAP of 65.7\% on M6Doc. These results further support the robustness and generalizability of our proposed method across diverse document analysis scenarios.

\begin{table}[htbp]
   \centering
      \caption{{\bf Comparisons of layout analysis methods on PubLayNet~\cite{zhong2019publaynet}, DocLayNet~\cite{DocLayNet}, and M6Doc~\cite{Cheng2023M6DocAL} datasets. }
      }
   \begin{adjustbox}{max width=0.46\textwidth}
   \begin{threeparttable}
     \centering
    
     \begin{tabular}{lccccc}
     \toprule
     \multirow{2}{*}{Methods} & \multicolumn{3}{c}{PubLayNet} & \multicolumn{1}{c}{DocLayNet} & \multicolumn{1}{c}{M6Doc}  \\
      \cmidrule(lr){2-4} \cmidrule(lr){5-5} \cmidrule(lr){6-6}
      &   P & R & F & mAP & mAP  \\
      \midrule 
Faster-RCNN~\cite{ren2015faster}             & 97.2     & 96.4     & 96.8    & 73.4      & 49.0  \\
Mask-RCNN~\cite{He2017MaskR}               & 94.0     & 95.5     & 94.7    & 73.5      & 40.1  \\
YOLOv5~\cite{glenn_jocher_2021_5563715}                  & 97.7     & 96.3     & 97.0    & 76.8      & 53.6  \\
        \midrule
        \ourmodel          & 98.0     & 96.8     & 97.4    & 77.3      & 65.7    \\
        
        \bottomrule
     \end{tabular}

   \end{threeparttable}
   \end{adjustbox}

   \label{table:more_layout_ana_res}
\end{table}

\begin{table*}[htbp]
\centering
\caption{Quantitative accuracy (\%) of cooperating SPOT prompting with Qwen2.5-VL-B using text spotting, key information extraction, table recognition, and layout analysis task datasets on OCRBench. Following TextMonkey~\cite{textmonkey}, we use the accuracy metrics to evaluate our method.}

\resizebox{0.97\linewidth}{!}{

\begin{tabular}{c|cc|ccc|ccc}
\toprule      \multirow{2}{*}{Method}
             & \multicolumn{2}{c|}{Scene Text-Centric VQA}        & \multicolumn{3}{c|}{Document-oriented VQA}                    & \multicolumn{3}{c}{KIE}   \\
              \cmidrule(lr){2-3}\cmidrule(lr){4-6} \cmidrule(lr){7-9} 
             & STVQA & TextVQA & DocVQA & InfoVQA & ChartQA & FUNSD   & SROIE  & POIE  \\ \midrule
BLIP2-OPT-6.7B~\cite{li2023blip2}   & 20.9  & 23.5  & 3.2    & 11.3           & 3.4                  & 0.2     & 0.1    & 0.3  \\
mPLUG-Owl~\cite{Ye2023mPLUGOwlME}    & 30.5  & 34.0  & 7.4    & 20.0             & 7.9           & 0.5     & 1.7    & 2.5   \\
InstructBLIP~\cite{Dai2023InstructBLIPTG} & 27.4  & 29.1   & 4.5    & 16.4           & 5.3             & 0.2     & 0.6    & 1.0     \\
LLaVAR~\cite{zhang2023llavar}       & 39.2  & 41.8   & 12.3   & 16.5           & 12.2              & 0.5     & 5.2    & 5.9    \\
BLIVA~\cite{Hu2023BLIVAAS}        & 32.1  & 33.3  & 5.8    & 23.6           & 8.7             & 0.2     & 0.7    & 2.1     \\
mPLUG-Owl2-8~\cite{Ye2023mPLUGOwI2RM}   & 49.8  & 53.9   & 17.9   & 18.9           & 19.4            & 1.4     & 3.2    & 9.9         \\
LLaVA1.5-7B~\cite{liu2024llavanextllava1.5}     & 38.1  & 38.7    & 8.5    & 14.7           & 9.3            & 0.2     & 1.7    & 2.5          \\
TGDoc~\cite{Wang2023TowardsIDtgdoc}   & 36.3 & 46.2  & 9.0           & 12.8                & 12.7       & 1.4    & 3.0   & 22.2   \\
UniDoc~\cite{feng2023unidoc}       & 35.2  & 46.2    & 7.7    & 14.7           & 10.9                    & 1.0       & 2.9    & 5.1    \\
DocPedia~\cite{feng2024docpedia}     & 45.5  & 60.2   & 47.1   & 15.2           & 46.9                 & {29.9}    & 21.4   & 39.9    \\
Monkey-8B~\cite{monkey}       & {54.7}  & 64.3    & {50.1}   & {25.8}           & {54.0}            & 24.1    & {41.9}   & 19.9  \\ 
InternVL-8B~\cite{Chen2023InternVLS}     & 62.2  & 59.8      &28.7    & 23.6           & 45.6  & 6.5    & 26.4    & 25.9   \\
InternLM-XComposer2-7B~\cite{Dong2024InternLMXComposer2MF}      & 59.6 & 62.2      &39.7    & 28.6           & 51.6  & 15.3   & 34.2    & 49.3   \\ 
TextMonkey-9B~\cite{textmonkey}   & 61.8 & 65.9    & 64.3          & 28.2                & 58.2       & 32.3    & 47.0   & 27.9   \\
InternVL2-2B~\cite{Chen2024HowFAinternvl2}   & 65.6 & 66.2    & 76.7          & 46.8                & {67.6}       & 42.0    & 68.0   & 66.8   \\
Mini-Monkey-2B~\cite{Huang2024MiniMonkeyMA} & {67.2} & {68.8} & {78.4} & {50.0} & 67.3 & {43.2} & {70.5} & {71.2} \\ 
Qwen2.5-VL-7B~\cite{Bai2025Qwen25VLTR} & {66.2} & {68.3} & {76.4} & {51.4} & {66.1 }& {45.9} & {70.4} & {70.5} \\ 
\midrule
Qwen2.5-VL-7B+SPOT & \textbf{68.6} & \textbf{69.8} & \textbf{78.9} & \textbf{52.8} & \textbf{67.7 }& \textbf{47.5} & \textbf{73.3} & \textbf{72.4} \\ 
\bottomrule
\end{tabular}}

\label{tab:supp_ocrbench_accuracy_metric}
\end{table*}

\subsection{Evaluation on FUNSD Dataset for KIE}
As shown in \cref{table:hallucination_of_res_model}, we have conducted additional key information extraction experiments on the FUNSD~\cite{Jaume2019FUNSDAD} dataset. As shown in Table E, our OmniParser V2 achieves a 1-NED score of 57.3\%, outperforming prior representative methods such as StrucTexT and LayoutLMv3. This further supports the effectiveness of our approach in diverse KIE scenarios.

\begin{table}[htbp]
   \centering
      \caption{{\bf Comparisons of key information extraction methods on FUNSD~\cite{Jaume2019FUNSDAD} dataset.} We present Normalized Edit Distance (1-NED) for the entity-level key information extraction. 
      }
   \begin{adjustbox}{max width=0.46\textwidth}
   \begin{threeparttable}
     \centering
    
     \begin{tabular}{lc}
     \toprule
     \multirow{1}{*}{Methods} & \multirow{1}{*}{ 1-NED }   \\
      
      \midrule 

       StrucTexT$_{Base}$   & 46.8\% \\
        LayoutLMv3$_{Base}$   & 53.1\% \\
        StrucTexTv2$_{Small}$ & 55.0\% \\
      \midrule 
        
        \ourmodel    & 57.3\% \\

        \bottomrule
     \end{tabular}

   \end{threeparttable}
   \end{adjustbox}

   \label{table:hallucination_of_res_model}
\end{table}

\subsection{Evaluation on OCRBench}
To further validate the generalizability of SPOT prompting in MLLMs, we evaluate Qwen2.5-VL-7B with and without SPOT prompting on OCRBench~\cite{Liu2023OCRBenchOT}, a comprehensive suite covering scene-text VQA, document VQA, and KIE. As shown in~\cref{tab:supp_ocrbench_accuracy_metric}, SPOT consistently improves accuracy across all tasks, including DocVQA (78.9\%), ChartQA (67.7\%), and SROIE (73.3\%), outperforming both baseline and other competitive MLLMs. These results confirm SPOT’s broad applicability in enhancing structured text understanding within MLLMs across diverse OCR benchmarks.

\subsection{Scalability to Diverse Document and Languages}
To evaluate scalability beyond standard English documents, we conduct layout analysis on the challenging M6Doc~\cite{Cheng2023M6DocAL} benchmark, which covers heterogeneous formats (scanned, photographed, and digital), varied layouts (multi-column, irregular blocks), and bilingual content (Chinese and English). As shown in~\cref{table:more_layout_ana_res}, \ourmodel achieves an mAP of \textbf{65.7\%} on M6Doc, significantly outperforming task-specific detectors such as Faster-RCNN (49.0\%), Mask-RCNN (40.1\%), and YOLOv5 (53.6\%). Notably, our method also performs competitively on PubLayNet and DocLayNet, suggesting its robustness across diverse document distributions. These results validate that the proposed framework is modular and language-agnostic, and effectively generalizes to complex real-world scenarios without requiring architecture modifications.

\subsection{Failure Case}
\begin{figure}[htbp]
\centering
\includegraphics[width=0.5\linewidth]{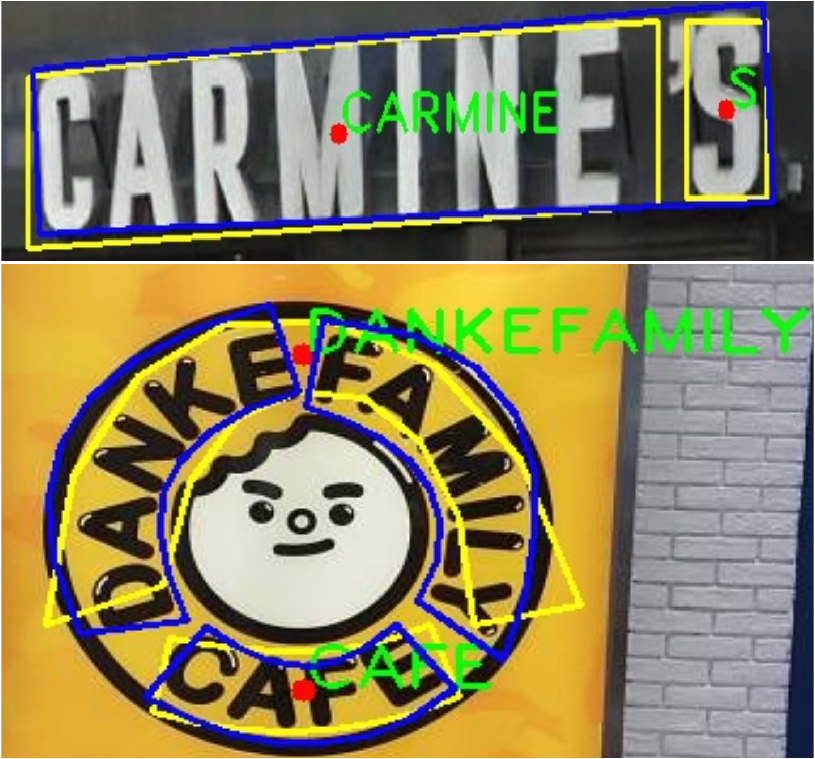}
\captionof{figure}{Failure cases, predicted \textcolor{red}{points} and \textcolor{deepyellow}{polygons}, \textcolor{blue}{GT polygons}.}
\label{fig:failure_case}
\end{figure}

Typical failure cases of inaccurately predicted points are demonstrated in \cref{fig:failure_case}. Our method only requires supervision of word locations in the training phase. 
It is quite robust to noisy location predictions.
Nonetheless, the accuracy of text spotting might be influenced when ambiguities arise in word point locations.

{
\bibliographystyle{IEEEtran}
 \bibliography{strings_abbr, references}

}